\documentclass[journal]{IEEEtran}
\usepackage{epsfig}
\usepackage{graphicx}
\usepackage{amsmath}
\usepackage{amssymb}
\usepackage{graphics} % for pdf, bitmapped graphics files
\usepackage{mathptmx} % assumes new font selection scheme installed
\usepackage{booktabs}
\usepackage{multirow}
\usepackage{color}
\usepackage{mathrsfs}
\usepackage{float}
\usepackage{subfigure}
\usepackage{xcolor}

\ifCLASSOPTIONcompsoc
  \usepackage[nocompress]{cite}
\else
  \usepackage{cite}
\fi
\ifCLASSINFOpdf
\else
\fi
\begin{document}
\title{Perception and Navigation in Autonomous Systems in the Era of Learning: A Survey}

\author{Yang~Tang$^{\dagger}$,~\IEEEmembership{Senior Member,~IEEE,}
        Chaoqiang~Zhao$^{\dagger}$,
        Jianrui~Wang,
        Chongzhen~Zhang,
        Qiyu~Sun,
        Wei~Xing~Zheng,~\IEEEmembership{Fellow,~IEEE,}
        Wenli~Du,
        Feng~Qian,
        and~J{\"u}rgen~Kurths% <-this % stops a space
\IEEEcompsocitemizethanks{\IEEEcompsocthanksitem This work was supported by National Natural Science Foundation of China (Basic Science Center Program: 61988101), National Natural Science Fund for Distinguished Young Scholars (61725301), the Programme of Introducing Talents of Discipline to Universities (the 111 Project) under Grant B17017, the Program of Shanghai Academic Research Leader (20XD1401300), CNPC Innovation Found (2021D002-0902) and Shanghai AI Lab.
\IEEEcompsocthanksitem $\dagger$ Y. Tang and C. Zhao contributed equally to this work.
\IEEEcompsocthanksitem Y. Tang, C. Zhao, J. Wang, C. Zhang, Q. Sun, W. Du and F. Qian are with the Key Laboratory of Smart Manufacturing in Energy Chemical Process, Ministry of Education, East China University of Science and Technology, Shanghai, 200237, China (e-mail: yangtang@ecust.edu.cn (Y. Tang)).
\IEEEcompsocthanksitem W. X. Zheng is with the School of Computer, Data and Mathematical Sciences,
Western Sydney University, Sydney, NSW 2751, Australia (e-mail: w.zheng@westernsydney.edu.au).
\IEEEcompsocthanksitem J. Kurths is with the Potsdam Institute for Climate Impact Research,
14473 Potsdam, Germany, with the Institute of Physics, Humboldt University
of Berlin, 12489 Berlin, Germany (e-mail: juergen.kurths@pik-potsdam.de).}
%\IEEEcompsocthanksitem Y. Tang and C. Zhao contributed equally to the work.% <-this % stops an unwanted space}
}
% note the % following the last \IEEEme

\markboth{IEEE Transactions on Neural Networks and Learning Systems}%
%\markboth{}%
{Shell \MakeLowercase{\textit{et al.}}: Bare Demo of IEEEtran.cls for Computer Society Journals}
\IEEEtitleabstractindextext{%
\begin{abstract}
Autonomous systems possess the features of inferring their own state, understanding their surroundings, and performing autonomous navigation.
With the applications of learning systems, like deep learning and reinforcement learning, the visual-based self-state estimation, environment perception and navigation capabilities of autonomous systems have been efficiently addressed, and many new learning-based algorithms have surfaced with respect to autonomous visual perception and navigation.
In this review, we focus on the applications of learning-based monocular approaches in ego-motion perception, environment perception and navigation in autonomous systems, which is different from previous reviews that discussed traditional methods. First, we delineate the shortcomings of existing classical visual simultaneous localization and mapping (vSLAM) solutions, which demonstrate the necessity to integrate deep learning techniques.
Second, we review the visual-based environmental perception and understanding methods based on deep learning, including deep learning-based monocular depth estimation, monocular ego-motion prediction, image enhancement, object detection, semantic segmentation, and their combinations with traditional vSLAM frameworks.
Then, we focus on the visual navigation based on learning systems, mainly including reinforcement learning and deep reinforcement learning. Finally, we examine several challenges and promising directions discussed and concluded in related research of learning systems in the era of computer science and robotics.
\end{abstract}

}

\maketitle

\IEEEdisplaynontitleabstractindextext

% \fi
\IEEEpeerreviewmaketitle

%\IEEEraisesectionheading{\section{Introduction}\label{sec:introduction}}]
\section{Introduction}

\begin{figure}[!t]
	\centering
	
	\includegraphics[width = \columnwidth]{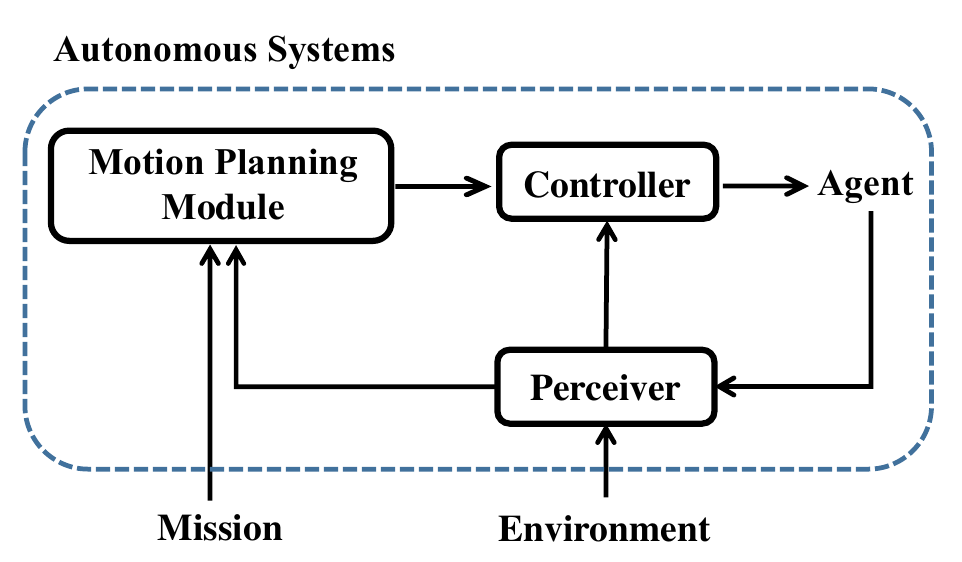}
	%\caption{fig1}
	\caption{An illustration for the autonomous systems.}
	\label{fig:fig1}
\end{figure}

\IEEEPARstart{I}{n} recent years, with the rapid developments in learning systems, such as deep learning and reinforcement learning, learning systems have been widely applied in various fields in smart grid \cite{quan2019survey}, biology\cite{mahmud2018applications}, finance \cite{deng2016deep}, object detection \cite{zhao2019object}, industrial production processes \cite{qian2017fundamental}, and particularly in the autonomous systems of robots.
Autonomous systems have gained a broad application prospect in various industries, including autonomous robots \cite{pfeiffer2017perception} and autonomous driving \cite{badue2019self,grigorescu2020survey}.
Although current autonomous systems can perform single, simple, and repetitive tasks, such as aided driving \cite{paden2016survey} and transportation \cite{loianno2017cooperative}, the future of autonomous systems has significant potential.
With the help of deep neural networks, autonomous systems that can learn and think like humans are becoming a reality.
Intelligent and autonomous systems are the ultimate aim, which can perform advanced tasks autonomously, interact with humans, and even work better than humans \cite{ingrand2017deliberation}. Primarily, the autonomy of autonomous vehicle systems relies on the results of in-depth environment perception, intelligent motion planning, and accurate control \cite{elfes1989using}. The architecture of autonomous systems is illustrated in Fig. \ref{fig:fig1}. Based on their perceiver \cite{cadena2016past,frazzoli2005maneuver}, autonomous systems understand their own state and surrounding environments by covering visual localization, mapping, and understanding the environment. Finally, autonomous systems can reach the designated position autonomously and complete advanced missions by combining the results of environment perception and motion planning with control signals.

Perceiving and understanding the environment are the basic elements of autonomous systems \cite{cadena2016past}. The development and application of visual simultaneous localization and mapping (vSLAM) have equipped robots with the ability to locate themselves and model the environment from vision, which has significantly expanded the autonomy and intelligence of robots. With the help of vSLAM, autonomous systems have the ability to use different visual sensors to collect environmental information, model their surroundings and estimate their current state \cite{sualeh2019simultaneous}.

\begin{figure}[t]
	\centering
	\subfigure[Sparse map;]{
		\includegraphics[width = 0.45\columnwidth]{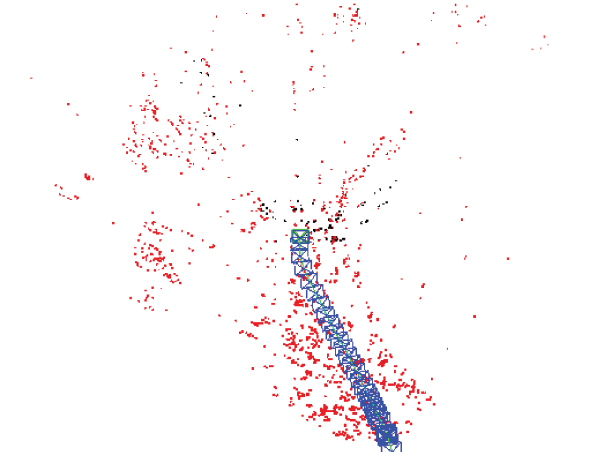}
		%\caption{fig1}
	}
	\subfigure[Semi-dense map;]{
		\includegraphics[width = 0.45\columnwidth]{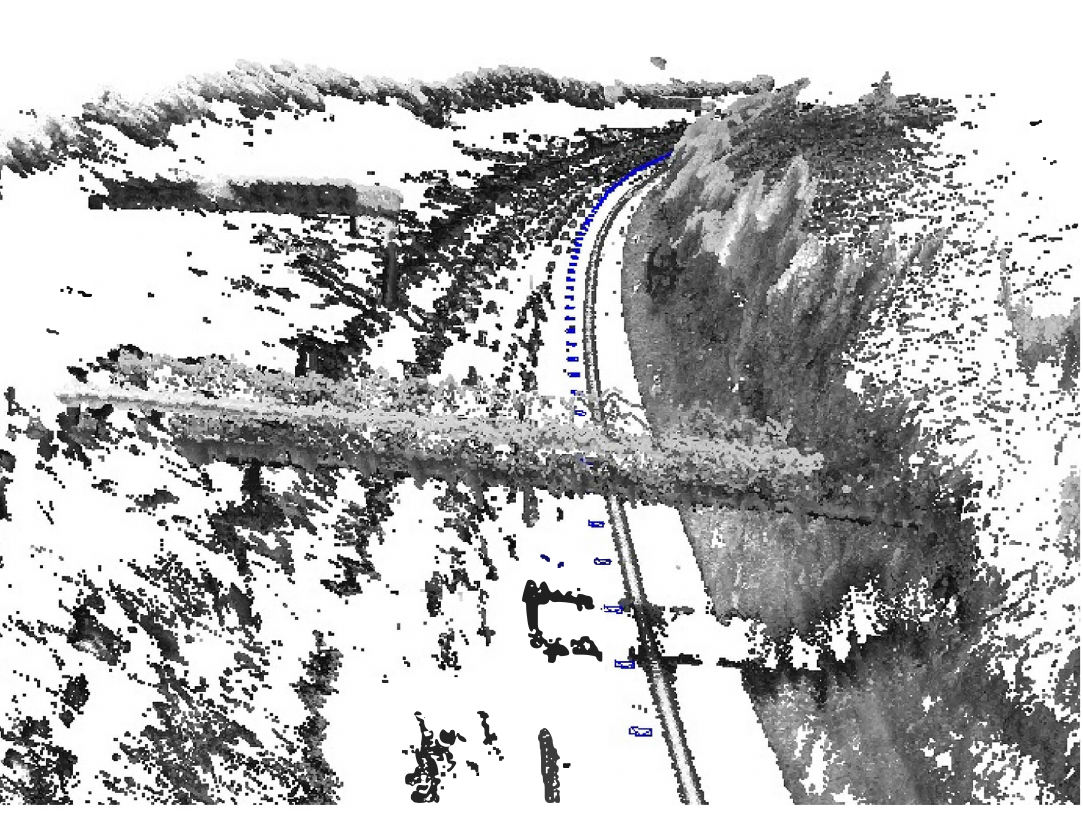}
	}
	\subfigure[Dense map;]{
		\includegraphics[width = 0.45\columnwidth]{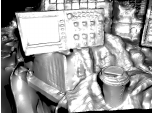}
		%\caption{fig1}
	}
	\subfigure[Semantic map.]{
		\includegraphics[width = 0.48\columnwidth]{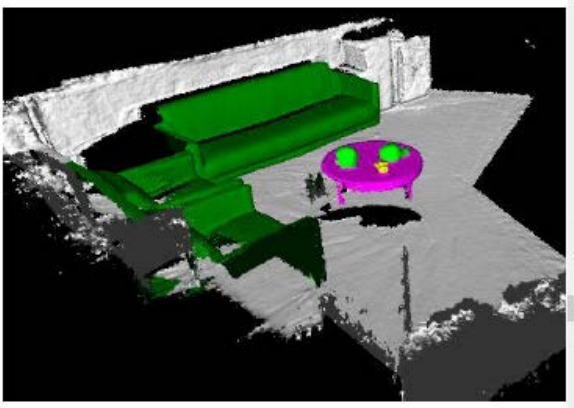}
	}
	\caption{The environments are represented in different types. (a): Sparse map is produced by ORB-SLAM2 \cite{mur2017orb}. (b): Semi-dense map is produced by LSD-SLAM \cite{engel2014lsd}. (c): Dense map is produced by DTAM \cite{newcombe2011dtam}. (d): Semantic map is produced by DA-RNN \cite{xiang2017rnn}.}
	
	\label{fig:fig3}
\end{figure}

\textbf{Perceiving the environment.} A good perception and understanding of the surrounding environment are indispensable for autonomous systems. vSLAM algorithms have benn widely applied to model the environments into different types based on the actual requirements, including sparse map \cite{mur2017orb}, semi-dense map \cite{engel2014lsd}, and dense map \cite{newcombe2011dtam}, as shown in Fig. \ref{fig:fig3} (a)-(c).
Although the geometric structures of surroundings in these representations are clearly perceived and modeled, a high-level information of these objects, like the semantic information, is still lacking.

\textbf{Perceiving their own state.} The state of an autonomous vehicle is described by its position and orientation. Understanding their current state in real time is important for autonomous systems, which is the main precondition of autonomous control.
Although current vSLAM algorithms play a crucial role in self-localization and ego-motion estimation, there are still some strong assumptions imposed in current vSLAM systems, such as the static scene hypothesis and the photometric consistency hypothesis.

\textbf{Visual navigation.} The ability of autonomous navigation is also essential in autonomous systems. When an autonomous vehicle is assigned a destination, it requires the capabilities of planning a reasonable path or trajectory. Poor or untimely planning may lead to terrible results, such as collision and crash. Therefore, the ability of human-like planning is the future development direction, and it is possible to achieve this intention with the help of learning framework. Since traditional motion planning methods have been well summarized in \cite{paden2016survey}, this review mainly focuses on the aspect of reinforcement learning-based navigation in autonomous systems.
%Trajectory planning algorithms aim to provide reasonable paths for vehicles to reach the destination, and these paths should avoid possible collisions with static or dynamic obstacles and satisfy the dynamics of the vehicles.

\textbf{Learning-based methods for visual perception and navigation.} With the development in learning framework \cite{pouyanfar2019survey}, deep learning and reinforcement learning have demonstrated outstanding performance in image processing \cite{frenay2013classification,druzhkov2016survey}, natural language processing \cite{sharma2017literature,lippi2019natural}, motion estimation \cite{kendall2015posenet}, game theory \cite{wang2022cooperative}, biology \cite{mahmud2018applications}, finance \cite{deng2016deep}, and control \cite{kiumarsi2017optimal}, etc.
The impact of learning framework on perception as well as navigation is transformational, and it has made significant advances in autonomous systems \cite{cadena2016past}.
Recently, deep learning-based models are widely used in relevant works of environment perception, such as monocular depth estimation \cite{zhou2017unsupervised}, ego-motion prediction \cite{kendall2015posenet}, objective detection \cite{zhao2019object}, and semantic segmentation \cite{ghosh2019understanding}. Furthermore, to improve the tracking, localization and mapping performance of current vSLAM methods in some complex environments (e.g., low light or night-time scenes), attempts have been made to incorporate vSLAM with deep learning and satisfactory results have been obtained \cite{jung2019multi}.
For example, some related works \cite{jadidi2017gaussian,gan2017sparse} incorporated learning based semantic understanding into the vSLAM to reconstruct the semantic maps of surroundings, as shown in Fig. \ref{fig:fig3} (d), thereby getting a high-level understanding of surroundings.
Moreover, related work in \cite{bruce2017one} has demonstrated that reinforcement learning exhibits good performance in robotic navigation. It resolved and implemented the navigation problems in an end-to-end manner. In addition, reinforcement learning enables robots to learn and imitate humans to make decisions. Unlike some well-written reviews \cite{fraundorfer2012visual, cadena2016past,sualeh2019simultaneous}, this survey mainly focuses on surveying the learning-based perception, including self-state perception and environment perception, as well as the representative results for reinforcement learning-based navigation in autonomous systems.

The rest of the paper is organized as follows: Section II introduces related works on visual perception, including a brief review of traditional vSLAM methods, deep learning-based visual perception, and methods combining deep learning with vSLAM. Section III provides an overview of the reinforcement learning-based visual navigation. Section IV summarizes the deficiencies and challenges of existing learning systems for visual perception and navigation, and provides some ideas about future directions. Finally, this survey is concluded in Section V.

\begin{table*}[]
	
	\scriptsize
	
	\centering
	
	\caption{A summary of major geometric vSLAM methods. ``Mono.'' denotes the monocular camera, and ``stereo'' stands for stereo camera. }
	
	\label{Tab01}
	\resizebox{2\columnwidth}{!}{
		\begin{tabular}{c|c|cc|ccc|ccc|c|}
			\toprule
			\multicolumn{1}{c}{}&\multicolumn{1}{c}{}&\multicolumn{2}{c}{Method}& \multicolumn{3}{c}{Type} & \multicolumn{3}{c}{Map}&\multicolumn{1}{c}{}\\
			\cmidrule(r){3-4}\cmidrule(r){5-7}\cmidrule(r){8-10}
			\hline
			Year  &Reference &Filtering-based &Optimization-based &Direct &Semi-direct &Feature-based & dense&semi-dense&sparse& Sensor \\
			\hline
			2003&Real-time SLAM \cite{davison2003real}&$\surd$&&&&$\surd$&&&$\surd$&Mono.\\
			2004&Davison \textit{et al.} \cite{davison2004real}&$\surd$&&&&$\surd$&&&$\surd$&Mono.\\
			2005&CV-SLAM \cite{jeong2005cv}&$\surd$&&&&$\surd$&&&$\surd$&Mono.\\
			2006&Smith \textit{et al.} \cite{smith2006real}&$\surd$&&&&$\surd$&&&$\surd$&Mono.\\
			2007&MonoSLAM \cite{davison2007monoslam}&$\surd$&&&&$\surd$&&&$\surd$&Mono.\\
			2007&PTAM \cite{klein2007parallel}&&$\surd$&&&$\surd$&&&$\surd$&Mono.\\
			2008&Silveira \textit{et al.} \cite{silveira2008efficient}&&$\surd$&$\surd$&&&&&$\surd$&Mono.\\
			2009&Migliore \textit{et al.} \cite{migliore2009use}&$\surd$&&&&$\surd$&&&$\surd$&Mono.\\
			2010&Newcombe \textit{et al.} \cite{newcombe2010live}&&$\surd$&&$\surd$&&$\surd$&&&Mono.\\
			2011&DTAM \cite{newcombe2011dtam}&&$\surd$&$\surd$&&&$\surd$&&&Mono.\\
			2011&Kinectfusion \cite{newcombe2011kinectfusion}&$\surd$&&$\surd$&&&$\surd$&&&RGB-D\\
			2012&Kintinuous \cite{kaess2012kintinuous}&&$\surd$&$\surd$&&&$\surd$&&&RGB-D\\
			2013&Weikersdorfer \textit{et al.} \cite{weikersdorfer2013simultaneous}&$\surd$&&$\surd$&&&&$\surd$&&Mono., Event camera\\
			2013&Endres \textit{et al.} \cite{endres20133}&&$\surd$&&&$\surd$&$\surd$&&&RGB-D\\
			2013&Li \textit{et al.} \cite{li2013high}&$\surd$&&&&$\surd$&&&$\surd$&Mono.,IMU\\
			2014&SVO \cite{forster2014svo}&&$\surd$&&&$\surd$&&$\surd$&&Mono.\\
			2014&LSD-SLAM \cite{engel2014lsd}&&$\surd$&$\surd$&&&&$\surd$&&Mono.\\
			2014&Weikersdorfer \textit{et al.} \cite{weikersdorfer2014event}&&$\surd$&&&$\surd$&&&$\surd$&RGB-D, Event camera\\
			2015&Stereo-LSD-SLAM \cite{engel2015large}&&$\surd$&$\surd$&&&&$\surd$&&Stereo\\
			2015&ORB-SLAM \cite{mur2015orb}&&$\surd$&&&$\surd$&&&$\surd$&Mono.\\
			2015&Leutenegger \textit{et al.} \cite{leutenegger2015keyframe}&&$\surd$&&&$\surd$&&&$\surd$&Stereo\\
			2015&Bloesch \textit{et al.} \cite{bloesch2015robust}&$\surd$&&$\surd$&&&&&$\surd$&Mono.,IMU\\
			2016&ElasticFusion \cite{whelan2016elasticfusion}&&$\surd$&$\surd$&&&$\surd$&&&RGB-D\\
			2016&Forster \textit{et al.} \cite{forster2016manifold}&&$\surd$&&&$\surd$&&&$\surd$&Mono., IMU\\
			2016&SVO 2.0\cite{forster2016svo}&&$\surd$&&$\surd$&&&&$\surd$&Mono., Multicamera\\
			2016&EVO \cite{rebecq2016evo}&&$\surd$&$\surd$&&&&$\surd$&&Event camera\\
			2017&DSO series \cite{engel2017direct,wang2017stereo,von2018direct}&&$\surd$&$\surd$&&&&&$\surd$&Mono., Stereo,IMU\\
			2017&ORB-SLAM2 \cite{mur2017orb}&&$\surd$&&&$\surd$&&&$\surd$&Mono., Stereo, RGB-D\\
			2017&Bundlefusion \cite{dai2017bundlefusion}&&$\surd$&&&$\surd$&$\surd$&&&RGB-D\\
			2017&Mur \textit{et al.} \cite{mur2017visual}&&$\surd$&&&$\surd$&&&$\surd$&Mono., IMU\\
			2018&ProSLAM \cite{schlegel2018proslam}&&$\surd$&&&$\surd$&&&$\surd$&Stereo\\
			2018&Sun \textit{et al.} \cite{sun2018robust}&$\surd$&&&&$\surd$&&&$\surd$&Stereo, IMU\\
			2018&ICE-BA \cite{liu2018ice}&&$\surd$&&&$\surd$&&&$\surd$&Mono., IMU\\
			2018&VINS-mono \cite{qin2018vins}&&$\surd$&&&$\surd$&&&$\surd$&Mono., IMU\\
			2018&Lee \textit{et al.} \cite{lee2018loosely}&&$\surd$&&$\surd$&&&&$\surd$&Mono.\\
			2019&BAD SLAM\cite{schops2019bad}&&$\surd$&$\surd$&&&$\surd$&&&RGB-D\\
			2019&RESLAM \cite{schenk2019reslam}&&$\surd$&&$\surd$&&$\surd$&&&RGB-D\\
			2020&Huang \textit{et al.} \cite{huang2020monocular}&&$\surd$&&$\surd$&&&&$\surd$&Mono.\\
			2021&OV$^{2}$SLAM \cite{ferrera2021ov}&&$\surd$&&$\surd$&&&&$\surd$&Stereo\\
			2021&ORB-SLAM3\cite{campos2020orb}&&$\surd$&&&$\surd$&&&$\surd$&Mono., Stereo, Fisheye, RGB-D,IMU\\
			
			\bottomrule
			
	\end{tabular}}
	
\end{table*}

%-------------------------------------------------------------------------
\section{Autonomous Visual Perception}

In autonomous systems, determining a comprehensive understanding of the environment and its current state is one of the basic and important perception tasks, which can be efficiently solved by vSLAM algorithms or sub-topics of vSLAM algorithms. Some classic SLAM methods are well summarized and discussed in \cite{fraundorfer2012visual,cadena2016past}. Cadena \textit{et al.} \cite{cadena2016past} reviewed the related works on SLAM over the last 30 years in detail. They revisited and answered several important and meaningful questions related to SLAM and stated that ``SLAM is necessary for autonomous robots". Different from previous review papers, in this section, we mainly focus on the application of deep learning algorithms in perception by subdividing them into three types.

\subsection{Geometric methods-based visual perception}

%\textbf{Traditional environment perception:}
SLAM is a common perception method in current autonomous systems.
Compared with the SLAM systems that use Lidar sensors \cite{montemerlo2002fastslam,montemerlo2003fastslam}, visual sensors such as RGB cameras \cite{forster2014svo,scaramuzza2011visual} can provide more environmental information, and they have been extensively investigated in recent years owing to their portability.
Therefore, we briefly summarize different types of vSLAM methods in a chronological order first, as presented in Table \ref{Tab01}. Their categories of optimization, maps, and sensors are enumerated in detail.
From Table \ref{Tab01}, we find that filtering-based vSLAM methods have been widely studied in the initial stage owing to their low computational burden. With the development in computer science, optimization-based vSLAM methods have become popular in recent years due to their higher accuracy. Meanwhile, dense maps are usually constructed by direct methods based on RGB-D sensors, like \cite{newcombe2011kinectfusion,kaess2012kintinuous,whelan2013robust}, $etc$. In addition, new sensors, such as event cameras, and multi-sensor data fusion are attracting significant attention and research prospects \cite{weikersdorfer2014event,bloesch2015robust,zhou2018semi,geneva2019efficient}.
In this section, we communicate the basic principles of the three classical monocular vSLAM solutions, including feature-based methods \cite{mur2015orb}, direct methods \cite{engel2014lsd}, and semi-direct methods \cite{forster2014svo}. The main difference between these three methods is the pose optimization by minimizing either the reprojection error, photometric error, or both \cite{scaramuzza2011visual}.

\textit{Feature-based methods} have dominated vSLAM for a long time, and different man-made features (like SIFT \cite{lowe2004distinctive}, SURF \cite{bay2006surf} and ORB \cite{rublee2011orb}) have been designed to improve their robust tracking and mapping in different scenarios. The feature-based methods can be divided into three parts, including image input, feature extraction and matching, tracking and mapping. Most recently, ORB-SLAM3 \cite{campos2020orb} is  proposed to support different kinds of sensors, like monocular, stereo, RGB-D, and IMU sensors, and it also supports a variety of camera models. ORB-SLAM3 system is much more versatile, accurate and robust than previous work.
However, the performance of feature-based methods relies on the correct matching, and they will fail to initialize and track in low-texture and repeated-texture scenes \cite{engel2014lsd} because of mismatch, suffers from the divergence in the optimization algorithm, and accumulation of drift.
\textit{Direct methods} cancel the process of feature extraction and matching, and the photometric information of pixels is directly used for pose and depth calculations during tracking and mapping \cite{engel2014lsd,gao2018ldso}. Direct methods regard the pose estimation as a nonlinear optimization problem and iteratively optimize the initial motion guess by minimizing the photometric error \cite{engel2014lsd}. Therefore, direct methods rely heavily on the luminosity consistency assumption \cite{von2020gn}.
\textit{Semi-direct methods} first establish feature correspondences on the basis of direct methods, which is the main difference from other methods \cite{forster2014svo,forster2016svo}. The principle of epipolar line constraint is applied to match the same features on the epipolar line. After matching the features, the solved pose is optimized by minimizing the reprojection error. Therefore, semi-direct methods handle the tracking problem by minimizing the photometric error and the reprojection error.
Similar to direct methods, semi-direct methods have a high requirement on image quality and are sensitive to photometric changes.

Although the architecture of vSLAM algorithms has been very maturely over the past 30 years and the three kinds of the above-mentioned approaches have achieved good performance in normal indoor/outdoor scenes, their tracking robustness and localization/re-localization accuracy in many complex scenarios (like high-dynamic/large scale/night-time environments or across weather/across-seasons conditions) still need to be further improved \cite{cadena2016past,pascoe2017nid,von2020gn}.

In conclusion, although traditional geometry-based vSLAM methods have achieved amazing results in environmental mapping and self-localization, these methods still have some shortcomings. For example, feature-based methods cannot adapt to low texture area; direct methods need a good initialization; semi-direct methods are sensitive to luminosity; traditional vSLAM/VO methods cannot handle changing lighting/weather/season conditions; monocular vSLAM/VO methods suffer from scale ambiguity and so on \cite{yousif2015overview,taketomi2017visual,von2020gn}. With the continuous development of deep learning in image processing, applying the latest deep learning systems to the existing vSLAM to handle the current problems in vSLAM methods is evolving into a popular research field.

\begin{table*}[]
	
	\scriptsize
	
	\centering
	
	\caption{A summary of deep learning-based monocular depth and ego-motion estimation. }
	
	\label{Tab02}
	\resizebox{2\columnwidth}{!}{
		\begin{tabular}{c|c|c|ccc|cc|c|}
			\toprule
			\multicolumn{1}{c}{}&\multicolumn{1}{c}{}&\multicolumn{1}{c}{}&\multicolumn{3}{c}{Supervisory signal}& \multicolumn{3}{c}{Mission} \\
			\cmidrule(r){4-6}\cmidrule(r){7-9}
			\hline
			Methods  &Years &Training Data &Supervised &Semi-supervised &Unsupervised &Depth &Pose  &Other tasks  \\
			\hline
			Eigen \textit{et al.} \cite{eigen2014depth} & 2014 &     RGB + Depth     &  $\surd$      &    &    & $\surd$ &          & -  \\
			Li \textit{et al.} \cite{li2015depth}       & 2015 &     RGB + Depth     &  $\surd$      &    &    & $\surd$ &          & Surface Normal \\
			Liu \textit{et al.} \cite{liu2015learning}  & 2015 &     RGB + Depth     &  $\surd$      &    &    & $\surd$ &          & - \\
			Mayer \textit{et al.} \cite{mayer2016large} & 2016 &     RGB + Depth     &  $\surd$      &    &    & $\surd$ &          & - \\
			Kendall \textit{et al.} \cite{kendall2017end} & 2017 &  Stereo images + Disparity     &  $\surd$      &    &    & $\surd$ &     & Disparity \\
			Fu \textit{et al.} \cite{fu2018deep}          & 2018 &     RGB + Depth     &  $\surd$      &    &    & $\surd$ &          & - \\
			Facil \textit{et al.} \cite{facil2019cam}     & 2019 &   RGB + Depth     &  $\surd$      &    &    & $\surd$ &          & - \\
			Huynh \textit{et al.} \cite{huynh2020guiding}     & 2020 &   RGB + Depth     &  $\surd$      &    &    & $\surd$ &          & - \\
			Ramamonjisoa \textit{et al.} \cite{ramamonjisoa2020predicting}     & 2020 &   RGB + Depth     &  $\surd$      &    &    & $\surd$ &     & Displacement field \\
			Chen \textit{et al.} \cite{chen2020improving}     & 2020 &   RGB + Depth     &  $\surd$      &    &    & $\surd$ &     & - \\
			Ranftl \textit{et al.} \cite{ranftl2020towards}     & 2020 &   RGB + Depth    &  $\surd$      &    &    & $\surd$ &     & - \\

			\hline
			Garg \textit{et al.} \cite{garg2016unsupervised}     & 2016 &   Stereo images   &   &  $\surd$  &   & $\surd$ &          & - \\
			Godard \textit{et al.} \cite{godard2017unsupervised} & 2017 &   Stereo images   &   &  $\surd$  &   & $\surd$ &          & - \\
			Kuznietsov \textit{et al.} \cite{kuznietsov2017semi} & 2017 &   Stereo images + LiDAR  &   &  $\surd$  &   & $\surd$ &          & -\\
			Poggi \textit{et al.} \cite{poggi2018learning}       & 2018 &   Stereo images   &   &  $\surd$  &   & $\surd$ &          & - \\
			Ramirez \textit{et al.} \cite{ramirez2018geometry}   & 2018 &   Stereo images + Semantic Label  &   &  $\surd$  &   & $\surd$ &          & - \\
			Aleotti \textit{et al.} \cite{aleotti2018generative} & 2018 &   Stereo images   &   &  $\surd$  &   & $\surd$ &          & - \\
			Pilzer \textit{et al.} \cite{pilzer2018unsupervised} & 2018 &   Stereo images   &   &  $\surd$  &   & $\surd$ &          & - \\
			Pilzer \textit{et al.} \cite{pilzer2019refine}       & 2019 &   Stereo images   &   &  $\surd$  &   & $\surd$ &          & - \\
			Tosi \textit{et al.} \cite{tosi2019learning}         & 2019 &   Stereo images   &   &  $\surd$  &   & $\surd$ &          & - \\
			Chen \textit{et al.} \cite{chen2019towards}          & 2019 &   Stereo images   &   &  $\surd$  &   & $\surd$ &          & Semantic segmentation \\
			Fei \textit{et al.} \cite{fei2019geo}                & 2019 &   Stereo images + IMU + Semantic Label  &   &  $\surd$  &   & $\surd$ &          & - \\
			Wang \textit{et al.} \cite{wang2018learning}                    & 2018 & Mono. sequences &  & $\surd$  &  & $\surd$ & $\surd$& - \\
			Zhan \textit{et al.} \cite{zhan2018unsupervised}                & 2018 & Stereo sequences&  & $\surd$  &  & $\surd$ & $\surd$& - \\
			Li \textit{et al.} \cite{li2018undeepvo}                        & 2018 & Stereo sequences&  &  $\surd$ &  & $\surd$ & $\surd$& - \\
			Wang \textit{et al.} \cite{wang2019unos}                        & 2019 & Stereo sequences&  &  $\surd$ &  & $\surd$ & $\surd$& Optical Flow \\
			Hur \textit{et al.} \cite{hur2020self}                          & 2020 & Stereo sequences&  &  $\surd$ &  & $\surd$ & $\surd$& Scene Flow \\
			Chi \textit{et al.} \cite{Chi2021CVPR}                       & 2021 & Stereo. sequences&  &  $\surd$ &  & $\surd$ & $\surd$& Optical Flow \\
			Jiao \textit{et al.} \cite{Jiao2021CVPR}                  & 2021 & Stereo. sequences&  &  $\surd$ &  & $\surd$ & $\surd$& Optical Flow, motion segmentation\\
			Jung \textit{et al.} \cite{jung2021fine}                    & 2021 & Mono. sequences &  & $\surd$  &  & $\surd$ & $\surd$& Semantic segmentation \\

			\hline

			Zhou \textit{et al.} \cite{zhou2017unsupervised}                & 2017 & Mono. sequences &  &   & $\surd$ & $\surd$ & $\surd$& Motion mask \\
			Vijayanarasimhan \textit{et al.} \cite{vijayanarasimhan2017sfm} & 2017 & Mono. sequences &  &   & $\surd$ & $\surd$ & $\surd$& Motion flow and segmentation\\
			Yang \textit{et al.} \cite{yang2017unsupervised}                & 2017 & Mono. sequences &  &   & $\surd$ & $\surd$ & $\surd$& Normal\\
			Mahjourian \textit{et al.} \cite{mahjourian2018unsupervised}    & 2018 & Mono. sequences &  &   & $\surd$ & $\surd$ & $\surd$& Principled Masks\\
			Zou \textit{et al.} \cite{zou2018df}                            & 2018 & Mono. sequences &  &   & $\surd$ & $\surd$ & $\surd$& Optical Flow \\
			Yin \textit{et al.} \cite{yin2018geonet}                        & 2018 & Mono. sequences &  &   & $\surd$ & $\surd$ & $\surd$& Optical Flow \\
			Ranjan \textit{et al.} \cite{ranjan2019competitive}             & 2019 & Mono. sequences &  &   & $\surd$ & $\surd$ & $\surd$& Optical Flow, Motion segmentation \\
			Wang \textit{et al.} \cite{wang2019unsupervised}                & 2019 & Mono. sequences &  &   & $\surd$ & $\surd$ & $\surd$& - \\
			Li \textit{et al.} \cite{li2019sequential}                      & 2019 & Mono.  sequences&  &   & $\surd$ & $\surd$ & $\surd$& - \\
			Li \textit{et al.} \cite{li2020self}                            & 2020 & Mono. sequences &  &   & $\surd$ & $\surd$ & $\surd$& - \\
			Vankadari \textit{et al.} \cite{vankadari2020unsupervised}      & 2020 & Mono. sequences&  &   & $\surd$ & $\surd$ & $\surd$& - \\
			Zhao \textit{et al.} \cite{Zhao2020masked}                & 2020 & Mono. sequences&  &   & $\surd$ & $\surd$ & $\surd$& - \\
			Spencer \textit{et al.} \cite{spencer2020defeat}                & 2020 & Mono. sequences&  &   & $\surd$ & $\surd$ & $\surd$& - \\
			Shu \textit{et al.} \cite{shu2020feature}                       & 2020 & Mono. sequences&  &   & $\surd$ & $\surd$ & $\surd$& Feature map \\
			Watson \textit{et al.} \cite{Watson2021CVPR}               & 2021 & Mono. sequences&  &   & $\surd$ & $\surd$ & $\surd$& - \\
			Lyu \textit{et al.} \cite{lyu2020hr}               & 2021 & Mono. sequences&  &   & $\surd$ & $\surd$ & $\surd$& - \\
			Zhou \textit{et al.} \cite{zhou_diffnet}               & 2021 & Mono. sequences&  &   & $\surd$ & $\surd$ & $\surd$& - \\

			\hline
			Konda \textit{et al.} \cite{konda2015learning}                  & 2015 & Mono. sequences + Pose  &  $\surd$  &      &       &      & $\surd$      & - \\
			Kendall \textit{et al.} \cite{kendall2015posenet}               & 2015 & Mono. sequences + Pose  &  $\surd$  &      &       &      & $\surd$      & - \\
			Costante \textit{et al.} \cite{costante2015exploring}           & 2015 & Mono. sequences + Pose  &  $\surd$  &      &       &      & $\surd$      & - \\
			Wang \textit{et al.} \cite{wang2017deepvo}                      & 2017 & Mono. sequences + Pose  &  $\surd$  &      &       &      & $\surd$      & - \\
			Xue \textit{et al.} \cite{xue2018guided}                        & 2018 & Mono. sequences + Pose  &  $\surd$  &      &       &      & $\surd$      & -\\
			Xue \textit{et al.} \cite{xue2019beyond}                        & 2019 & Mono. sequences + Pose  &  $\surd$  &      &       &      & $\surd$      & -\\
			Clark \textit{et al.} \cite{clark2017vinet}                     & 2017 & Mono. sequences + Pose + IMU &  $\surd$  &      &       &      & $\surd$      & -\\
			Chen \textit{et al.} \cite{chen2019selective}                   & 2019 & Mono. sequences + Pose + IMU &  $\surd$  &      &       &      & $\surd$      & - \\
			Xue \textit{et al.} \cite{xue2020learning}                      & 2020 & Mono. sequences + Pose  &  $\surd$  &      &       &      & $\surd$      & - \\
			Wei \textit{et al.} \cite{wei2020deepsfm}                      & 2020 & Mono. sequences + Pose + Depth &  $\surd$  &      &       &      & $\surd$      & - \\
			Zhang \textit{et al.} \cite{Zhuang2021CVPR}                    & 2021 & Mono. sequences + Pose  &  $\surd$  &      &       &      & $\surd$      & - \\%Zhuang2021CVPR
			\bottomrule
			
	\end{tabular}}
	
\end{table*}

\subsection{Deep learning-based visual perception}

With the development in deep learning, utilizing deep neural networks to address computer vision tasks has evolved into a popular research field in recent years. Many sub-topics of vSLAM for environment perception have been extensively studied based on deep learning, such as monocular depth and ego-motion prediction, which will be specified in the following sections.

\subsubsection{Learning-based monocular depth perception}
Depth is one of the most important information for autonomous systems in scene reconstruction, self-localization, obstacle avoidance and so on. Although the active depth sensors are available for depth perception, image-based techniques are often preferred thanks to the increasing availability of standard cameras on most consumer devices \cite{poggi2021confidence}.
Structure-from-motion (SfM) \cite{hartley2003multiple,furukawa2015multi } and stereo matching \cite{scharstein2002taxonomy,hirschmuller2005accurate} are two of the most popular methods to recover the depth from sequential or left and right images \cite{scharstein2002taxonomy}, and the depth is calculated by the triangulation and continuously optimized by projection cost and matching cost. However, the above methods rely on the assumption that multiple observation of the scene are available \cite{garg2016unsupervised}, which means that the above methods are not well applicable to estimating depth from a single image.

Estimating the depth from a single image is an ill-posed problem \cite{eigen2014depth}, which requires significant man-made prior knowledge when handled by traditional geometric methods \cite{karsch2012depth,ladicky2014pulling}.
Deep neural networks can recover pixel level depth information from single images in an end-to-end manner based on the prior knowledge learnt from ground truth depth labels or geometric relationships between images \cite{zhao2020monocular}. Since both ground-truth based supervised methods and geometry based unsupervised
methods have been well summarized in \cite{zhao2020monocular}, in this paper, we will focus on the latest work, starting with issues that remain unresolved in monocular depth estimation. As shown in \cite{zhao2020monocular}, monocular depth estimation has made great progress in recent few years, and unsupervised methods are already close to that of supervised methods. After several years of development, the framework of monocular depth estimation has become very mature. Recent work focuses on improving the deficiencies of the existing unsupervised framework, like static scenario assumptions \cite{kumar2021syndistnet} and photometric consistency assumptions \cite{spencer2020defeat}.

The smoothness loss is one of the most widely used constraints \cite{zhou2017unsupervised,yin2018geonet,ranjan2019competitive} to promote the smoothness of the surface depth of the object, for example, the depth of the adjacent points on the road surface varies by gradient. However, the existing methods do not impose smoothness constraints after distinguishing different targets in the scenario, resulting in smoothing edge areas that should be sharp in the estimated depth map. To address this problem, Yin \textit{et al.} \cite{yin2019enforcing} proposed a novel geometric constraint to improve the accuracy of depth estimation as well as the geometric shape in the predicted depth map by considering the surface normal. Instead of using additional constraints to get a clear geometric structure in monocular depth estimation, the method proposed in \cite{ramamonjisoa2020predicting} predicted a 2D displacement field of the given depth map to re-sample pixels around the occlusion boundaries into sharper reconstructions.
A recent study \cite{Watson2021CVPR} showed that incorporating sequence information into monocular framework is helpful to improve depth prediction, when the sequence information is available. Instead of estimating the accurate depth of each pixel, predicting the relative depth of pixels in the image is also crucial for scene perception and understanding \cite{Lienen2021CVPR}, which can also obtain good results in recovering metric depth. When considering the widely application scenarios of high-resolution depth maps, like object detection and semantic segmentation, Miangoleh \textit{et al.} \cite{Miangoleh2021CVPR} proposed to infer high-resolution depth maps from images based on pre-trained depth models.

Since the supervised signal of unsupervised methods is mainly based on the view reconstruction loss \cite{zhou2017unsupervised}, view reconstruction relies heavily on static scenario assumptions. Therefore, these methods fail to predict depth for moving objects. To deal with this challenge, Godard \textit{et al.} \cite{godard2019digging} designed an ``Auto-Masking'' to selectively eliminate pixels that keep the same position with same RGB value between adjacent frames in the sequence.
However, this method can only eliminate the influence of objects moving at equivalent relative translation to the camera, while other dynamic objects will still have negative influence on the unsupervised training process. Therefore, with the help of semantic segmentation, Klingner \textit{et al.} \cite{klingner2020self} divided the dynamic and static objects by the correspondence of class labels between frames, which is calculated by projection. Then, they eliminated the effects of these dynamic regions on view reconstruction loss.

There are also some novel studies that improve the accuracy of monocular depth estimation by utilizing the novel network framework, such as proposing novel depth network \cite{guizilini20203d} or using novel attention mechanism \cite{huynh2020guiding,johnston2020self}. Introducing traditional geometry is also a good way, Wang \textit{et al.} \cite{wang2018learning} tried to get a better pose estimation by using direct methods during training. The direct method was used to further optimize the output of the pose network before training, thereby getting a more accurate pose and depth estimation. Depth estimation based on novel cameras, like event-based camera \cite{zhu2018multivehicle}, fisheye camera \cite{kumar2021syndistnet} and panorama camera \cite{jiang2021unifuse}, is attracting increase attention because of its advantages, like low latency and wide field-of-view. Inspired by the high performance of HRNet \cite{SunXLW19}, Zhou \textit{et al.} \cite{zhou_diffnet} introduced the HRNet into the unsupervised monocular depth estimation task and got a satisfactory results.

Monocular depth estimation in special scenarios, such as adverse weather conditions and night-time scenes, is gradually being focused. Because of the complex luminosity changes and photometric inconsistency at night, the previous unsupervised frameworks driven by view reconstruction consistency \cite{zhou2017unsupervised,yin2018geonet,ranjan2019competitive} cannot be applied to the night-time scene \cite{spencer2020defeat} directly. Recent studies have tried to address this problem by using warped feature consistency \cite{spencer2020defeat} or cross-domain feature adaptation \cite{vankadari2020unsupervised}, which achieved good accuracy in night-time depth estimation. Spencer \textit{et al.} \cite{spencer2020defeat} designed a DeFeat-Net to simultaneously learn the cross-domain dense feature representations of frames. Moreover, a robust feature reconstruction consistency instead of view reconstruction consistency is used as the main supervised signal for the training of framework, thereby being able to adapt special scenarios. Based on the auto-encoder depth network pre-trained on day time, Vankadari \textit{et al.} \cite{vankadari2020unsupervised} used an additional night-time encoder to encode the images of night time. A PatchGAN-based adversarial discriminator was designed to constrain the consistency between the features among the images of day time and night time, which are encoded by two encoders respectively. Hence, the pre-trained decoder can directly recover a depth map of a night-time image from features encoded by the night-time encoder. Zhao \textit{et al.} \cite{zhao2021unsupervised} proposed to use a cyclegan-based domain adaptation framework to get an end-to-end night-time depth model from a pretrained day-time model, and it got a better results in night and even rainy night. Instead of using adaptation methods, Wang \textit{et al.} \cite{wang2021regularizing} leveraged a Mapping-Consistent Image Enhancement module to deal with the low visibility and a Statistics-Based Mask (SBM) to tackle textureless regions, so their work can directly train the model on night-time image sequences.

\subsubsection{Learning-based monocular ego-motion perception}
Visual odometry (VO) is the process of estimating the ego-motion of an agent (e.g., vehicle, human, and robot) by using the input of a single or multiple attached cameras \cite{fraundorfer2012visual}.
Geometry-based monocular VO methods handle the localization and tracking by minimizing the photometric error \cite{engel2017direct} or reprojection error \cite{mur2015orb} on sequential images. The difference between traditional VO and vSLAM is that VO system lacks the loop-closure detection and global optimization \cite{fraundorfer2012visual}. With the development of deep learning systems, using the features extracted by deep neural networks to regress the ego-motion in an end-to-end manner is becoming a hot application in recent years \cite{kendall2015posenet}. Compared with traditional VO methods, pose networks do not require complex parameter tuning, such as the settings of key frames and features \cite{kendall2015posenet}. Moreover, pose networks can learn the scale information from the ground truth during training, so these methods solve the monocular scale ambiguity problem that widely existed in traditional monocular VO methods \cite{engel2017direct,mur2015orb}.
Konda \textit{et al.} \cite{konda2015learning} first estimated the motion information through deep learning-based methods by formulating pose prediction as a classification problem. Alex \textit{et al.} \cite{kendall2015posenet} first demonstrated the ability of convolutional neural networks (CNNs) on 6-DOF pose regression. A deep CNN framework called PoseNet was designed for regressing monocular camera pose that could operate in different scenes in real-time. In \cite{costante2015exploring}, Costante \textit{et al.} also used a deep CNN to learn high-level feature representation, and the major difference from \cite{kendall2015posenet} is that the dense optical flow was calculated and used to estimate the ego-motion instead of feeding RGB images into the CNN directly. Considering the dynamics and relations between adjacent pose transformations, Wang \textit{et al.} \cite{wang2017deepvo} and Xue \textit{et al.} \cite{xue2018guided} used recurrent neural networks (RNNs) for camera localization. Then, Xue \textit{et al.} \cite{xue2019beyond} further extended their work by incorporating two helpful modules named ``Memory" and ``Refining" into VO tasks, which outperformed the previous deep learning-based VO methods \cite{xue2018guided}.

As the learning system is constantly evolving, introducing new learning architecture to current tasks has been a good way to improve the ability of pose network in high-level feature extraction and pose regression. Xue \textit{et al.} \cite{xue2020learning} proposed to construct a view graph to excavate the information of the whole given sequence for absolute camera pose estimation, and a graph neural network was applied to model the total graph. Li \textit{et al.} \cite{li2020self} introduced online meta-learning algorithms into previous learning framework, so that their method can continuously adapt to unseen environments in a self-supervised manner. Considering the error accumulation problem commonly suffered by previous learning-based methods, Zou \textit{et al.} \cite{zou2020learning} tried to aggregate long-term temporal information by using Conv-LSTM (convolutional long short term memory) to model long-term temporal dependency. Meanwhile, long-range constraints based on long-range image snippets are used to improve temporal consistency over long sequences, just like the local optimization (bundle adjustment) that widely used in traditional VO methods. Chi \textit{et al.} \cite{Chi2021CVPR} studied the performance difference between feature-level collaboration and loss-level joint optimization for multi-task learning (depth, pose and optical flow), and feature-level collaboration shows much greater performance improvement for all three tasks. Therefore, they designed a single network to integrate all the three tasks, and the pose component regresses pose from both images and estimated disparity map and optical flow. Inspired by bundle adjustment, Wei \cite{wei2020deepsfm} proposed a deep learning framework that iteratively improves both depth and pose based on the cost volume explicitly built to measure photo-consistency and geometric-consistency.
Zhuang \textit{et al.} \cite{Zhuang2021CVPR} presented an uncertainty based probabilistic framework that integrating pose predictions from deep neural networks and solutions from geometric feature-based solvers (5-point method and bundle adjustment).
Instead of estimating poses from images, Zhao \textit{et al.} \cite{zhao2020towards} recovered relative pose by directly solving the fundamental matrix from dense optical flow correspondence, which was predicted by an optical flow network, and the results demonstrated the effectiveness of the framework in pose estimation. Jiao \textit{et al.} \cite{Jiao2021CVPR} obtained the pose between frames by minimizing the reprojection error, since the optical flow and depth are predicted by deep neural networks.

The traditional methods have proved that combining visual information with inertial information is helpful for improving the visual localization accuracy \cite{jones2011visual,leutenegger2015keyframe,li2017monocular}. However, these visual-inertial odometry (VIO) methods suffer from accurate calibration between sensors, time-stamp synchronization between inertial and visual data, and effective inertial and visual information fusion  \cite{jones2011visual,leutenegger2015keyframe,li2017monocular}. Researchers believe that inertial information is also helpful in learning-based methods. Therefore, Clark \textit{et al.} \cite{clark2017vinet} proposed the first end-to-end VIO framework based on deep learning without the need for time-stamp alignment and manual calibration between different sensors. They used the CNN architecture to extract visual features and long short-term memory (LSTM) to extract the inertial features, and fused their features using a core LSTM processing module for pose regression. For a better integration of visual and inertial features extracted by the deep neural networks, Chen \textit{et al.} \cite{chen2019selective} presented a selective sensor fusion framework based on the attention mechanism, which autonomously selects the most useful features extracted from images or inertial measurement unit (IMU) by deep neural network. Therefore, even in the case of poor image quality, their algorithm can get accurate poses with the help of inertial data.

We briefly summarize the deep learning-based monocular depth and ego-motion estimation according to their published years, the training data, the training mode, and the missions, as shown in Table \ref{Tab02}. From the table, we find that attention has been paid increasingly to unsupervised methods these years, because unsupervised methods do not require expensive ground truth \cite{zhou2017unsupervised}. Besides, considering the in-depth relationship between projection and optical flow between frames, researchers always extend the unsupervised pose and depth estimation framework with optical flow estimation\cite{zou2018df,yin2018geonet}. Recently, scene flow (optical flow in 3D space) estimation \cite{Jiao2021CVPR,hur2020self} is getting more attention, which is trained together with depth and pose network in an unsupervised manner. Since optical flow, scene flow, depth and pose are tightly coupled, the training strategy will have an impact on the performance of each task \cite{Jiao2021CVPR}. Therefore, the multi-task frameworks have become popular in recent years, and the geometric relationship between these tasks (flow, segmentation, mask) has been exploited to improve the performance of the network. Besides, the data from multi-sensors (like camera, IMU) has also been added to join the network to provide additional information \cite{chen2019selective,fei2019geo}, thus promoting the training of networks.

\subsection{Deep learning with vSLAM}

The methods combining vSLAM with deep learning have also been extensively studied and have lead to notable improvements to traditional vSLAM methods, like tackling the scale ambiguity of monocular vSLAM \cite{tateno2017cnn,sucar2018bayesian}, improving the robust tracking and accurate mapping of vSLAM \cite{yang2018deep,zhao2019deep}, strengthening the adaptability of vSLAM in different environments \cite{jung2019multi, wang2019computationally,von2020gn}, and extending the semantical perception of the environments \cite{cheng2017dense,mccormac2017semanticfusion}.

\subsubsection{Learning-based monocular depth estimation and vSLAM}

Depth information plays an important role in traditional vSLAM methods, and sensor-based and triangulation-based methods are two basic ways to obtain the depth of features.
With the development of deep learning in the field of monocular depth estimation, researchers are trying to use deep learning-based methods as an alternative to the traditional depth calculation methods of vSLAM.
The combination of deep learning-based depth estimation and traditional vSLAM methods has been proved to be effective in obtaining the depth of features and overcoming the monocular scale ambiguity, thereby improving mapping and replacing the RGB-D sensors \cite{laina2016deeper,tateno2017cnn}.
Depth prediction was first introduced to dense monocular vSLAM by Laina \textit{et al.} \cite{laina2016deeper}. Since the mapping process reduces the dependence on feature extraction and matching, this method has the potential to reconstruct low-texture scenes. Moreover, this work showed that the depth estimation network can replace the depth sensors (such as RGB-D) and can be used for dense reconstruction. After that, a real-time dense vSLAM framework was proposed in \cite{tateno2017cnn}. They used the LSD-SLAM \cite{engel2014lsd} as the baseline and fused the depth estimation and semantic information. Unlike the work by Laina \textit{et al.} \cite{laina2016deeper}, where the depth estimation was directly used in vSLAM, Tateno \textit{et al.} \cite{tateno2017cnn} considered the predicted depth map as the initial guess of LSD-SLAM, and further refined the predicted depth value by the local or global optimization algorithms in vSLAM. This method not only got a higher pose accuracy than LSD-SLAM, but also overcame the issue of scale inconsistency in dense monocular reconstruction. Similarly, Yang \textit{et al.} \cite{yang2018deep} proposed a novel semi-supervised disparity estimation network and incorporated it into direct sparse odometry (DSO) \cite{eigen2014depth}, thereby achieving a better accuracy to monocular DSO and attaining a comparable performance to previous stereo DSO methods. Recently, Loo \textit{et al.} \cite{loo2019cnn} presented a CNN-SVO pipeline that leveraging the SVO \cite{forster2014svo} with depth prediction network to improve the mapping and tracking of SVO. Czarnowski \textit{et al.} \cite{czarnowski2020deepfactors} proposed a real-time probabilistic dense vSLAM system that integrates learned priors (depth) over geometry with classical vSLAM formulations in a probabilistic factor-graph formulation, and got a better accuracy than \cite{tateno2017cnn} in both trajectory and depth estimation. Combining depth estimation with vSLAM has been proven to effectively improve the performance of traditional monocular vSLAM. Moreover, vSLAM can also be used to promote the accuracy of depth networks.
For example, Tawari \textit{et al.} \cite{tiwari2020pseudo} proposed a self-improving framework. On the one hand the predicted depth was used to perform RGB-D feature-based vSLAM. On the other hand, the pose calculated by RGB-D feature-based vSLAM instead of that predicted by pose network was leveraged to train the depth network, thereby leading to more accurate depth estimation. The above works have shown how to integrate vSLAM with depth prediction via a deep neural network, and it is a promising direction to address inherent limitations of traditional vSLAM, especially with respect to estimating the absolute scale and obtaining dense depths.

\subsubsection{Learning-based pose estimation and vSLAM}

Although pose networks have achieved real-time performance and satisfactory accuracy, the existing learning based pose estimation methods do not include the mapping thread \cite{kendall2015posenet,wang2017deepvo}, which is important for the perception of the environmental structure. Besides, traditional direct methods rely heavily on the initial guess of pose during tracking, resulting in instable initialization and inaccurate tracking \cite{eigen2014depth,von2020gn}.
Therefore, combining learning-based pose estimation with traditional vSLAM is a good way to overcome the above deficiencies \cite{zhao2019deep,yang2020d3vo}.
Zhao \textit{et al.} \cite{zhao2019deep} designed a self-supervised pose prediction network and incorporated it into DSO \cite{eigen2014depth}. They considered the output of the pose network as the initial pose guess of direct VO, which replaced the constant motion model used in DSO; then, the initial pose was improved by the nonlinear optimization in DSO. This method got a more robust initialization and tracking than traditional DSO when testing on the KITTI odometry sequences \cite{geiger2013vision}. Yang \textit{et al.} \cite{yang2020d3vo} also focused on this field, and they proposed a novel framework for monocular VO that exploits deep networks on three levels - deep depth, pose and uncertainty estimation, which not only improve the robust initialization and tracking of DSO in the challenging scenarios with photometric changes but also assist in recovering the scale information of monocular VO.
Different from the above frameworks, Wagstaff \textit{et al.} \cite{wagstaff2020self} proposed to use a deep neural network to correct the pose estimated by traditional VO frameworks, and a self-supervised deep pose correction network is designed to estimate a pose correction rather than the full inter-frame pose. Teed \textit{et al.} \cite{teed2021droid} proposed a new deep learning-based SLAM system with strong performance and generalization, called DROID-SLAM, and a GRU based update operator is proposed for depth and pose update.

\subsubsection{Learning-based image enhancement and vSLAM}

Current monocular vSLAM methods have achieved good robustness under specific scenarios, such as outdoor sunny scenes with normal illumination conditions \cite{geiger2013vision}. While driving in complex environments, such as during the night, in rain, and other scenarios, current monocular vSLAM systems cannot accurately estimate the pose of robots and reconstruct the point clouds of environments.For example, compared with the day-time scenarios under a single light source (sun), the night-time scenarios suffer from complex lighting changes because of multiple light sources (e.g. street lights, own car light and other car lights) \cite{vankadari2020unsupervised}, which affects the extraction of high-quality points and the accuracy of feature matching between frames \cite{jung2019multi}. Therefore, tracking the key points or features between images in such scenarios is instable and in accuracy because of the changing illumination, resulting in inaccurate calculation of depth and pose \cite{jung2019multi}.
Besides, different weather conditions in the same scenario also cause changes in the luminosity and feature descriptors of the scene, which brings challenges to feature matching and relocalization \cite{pascoe2017nid}.
%Therefore, tracking in poor illumination conditions is still a big challenge for current SLAM systems .
Researchers have proposed some geometric methods to improve the performance of vSLAM in challenging environments, such as using multiple cameras \cite{liu2018towards}, designing a new NID metric \cite{pascoe2017nid} and raising novel feature descriptor \cite{alismail2016direct}, and these methods can achieve robust VO and relocalization under different lighting and weather conditions.

With the development of learning systems in image style translation \cite{zhu2017unpaired,isola2017image} and video synthesis \cite{chen2017coherent,gao2018reconet}, deep learning-based image enhancement provides a new and simple way for vSLAM systems to overcome challenging environments \cite{guo2016lime,anoosheh2019night}.
Deep learning-based image enhancement helps to enhance the quality of images, like enhancing the brightness constancy of images \cite{gomez2018learning} or transferring the images from low light to normal light \cite{jung2019multi}, so as to make the images more suitable for current vSLAM systems.
Considering that direct methods cannot handle the dynamic lighting changes, Gomez \textit{et al.} \cite{gomez2018learning} used deep neural networks to enhance the brightness constancy of image sequences captured from high dynamic range (HDR) environments. The experiments showed that learning-based image enhancement can improve the trajectory estimation of ORB-SLAM \cite{mur2015orb} and DSO \cite{eigen2014depth} in HDR environments. Since the illumination influences feature extraction and matching, Jung \textit{et al.} \cite{jung2019multi} proposed a new framework called multi-frame GAN that translated the images from night-time to day-time to improve the quality of input images. Both stereo ORB-SLAM \cite{mur2017orb} and stereo DSO \cite{wang2017stereo} achieved accurate tracking performance on the transferred high-quality day-time images, which means that their method \cite{jung2019multi} can overcome the low light environments. Unlike \cite{jung2019multi}, Von Stumberg \textit{et al.} \cite{von2020gn} replaced the input of direct methods (gray-scale images) with feature maps created by their designed GN-Net for relocalization tracking. Since GN-Net can predict the consistent feature map of the same scene under different lighting and weather conditions, their method had the ability to achieve accurate tracking and relocalization in different weather conditions.

\begin{figure}[t]
	\centering
	\subfigure[The trajectory generated by ORB-SLAM2 \cite{mur2017orb} in high-dynamic environment;]{
		\includegraphics[width = 0.5\columnwidth]{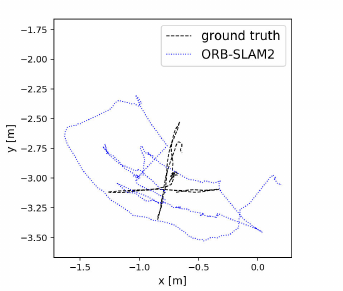}
		%\caption{fig1}
	}
	\subfigure[The trajectory generated by deep learning methods in a normal environment;]{
		\includegraphics[width = 0.4\columnwidth]{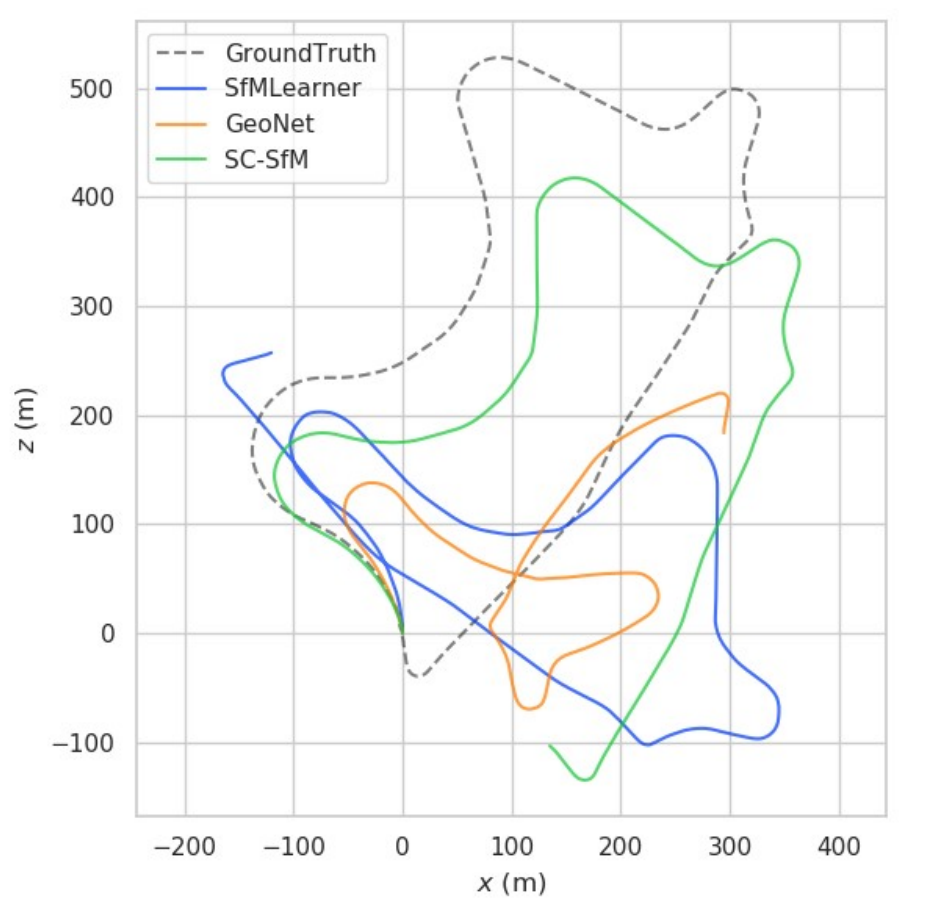}
	}
	\caption{(a): ORB-SLAM2 cannot generate the usable trajectory in high-dynamic environment \cite{cui2019sof}; (b): The trajectories generated by SfMLearner \cite{zhou2017unsupervised},GeoNet \cite{yin2018geonet} and SC-SfM \cite{bian2019unsupervised} on the KITTI odometry sequence 09.}
	
	\label{fig:fig2}
\end{figure}

\subsubsection{Learning-based object detection, semantic segmentation and vSLAM}

We consider the following three problems:

\textbf{Dynamic scene adaptability:} Traditional vSLAM relies heavily on static scene assumption, i.e., the performance of vSLAM is limited by moving objects \cite{cui2019sof}, as shown in Fig. \ref{fig:fig2} (a). Both photometric error and reprojection error are based on geometric projections between frames, but the features on the dynamic object do not satisfy the projection relationship based on the camera motion, which will lead to inaccurate pose estimation.
The features on static objects are positive to improve the accuracy, while those on dynamic objects have a negative impact on the tracking process \cite{xiao2019dynamic}. Therefore, if the dynamic objects on the input images can be detected and labelled, this problem will be well addressed. Considering the outstanding performance of deep learning-based object detection and semantic segmentation, the integration of deep learning framework and vSLAM can effectively assist vSLAM in identifying dynamic objects in the environment to classify and handle the dynamic features. Excellent detection and segmentation networks, such as YOLO \cite{redmon2016you}, SSD \cite{liu2016ssd}, Mask-RCNN, \cite{he2017mask} and SegNet \cite{badrinarayanan2017segnet}, have been incorporated into traditional vSLAM frameworks as an additional thread to identify and eliminate the dynamic features. Zhong \textit{et al.} \cite{zhong2018detect} presented a novel system that integrated vSLAM with the object detector SSD, called Detect-SLAM. The SSD was used to detect the dynamic and static objects for every key frame; Since the extracted features on the dynamic objects were removed, the remaining static features satisfied the projection function between frames, which greatly improved the accuracy of the pose and depth solutions.
Wang \textit{et al.} \cite{wang2019computationally} considered the effects of moving objects on localization accuracy and constructed maps, and developed a novel vSLAM solution. They used YOLOv3 \cite{redmon2018yolov3} to detect moving objects and constructed a semantic static map with the data without moving objects. Xiao \textit{et al.} \cite{xiao2019dynamic} developed a new detection thread to detect and remove the dynamic objects, and designed a selective tracking algorithm to process the dynamic features during tracking. Because the object detection methods are not considered during pixel-level semantic annotation, the classification of feature attributes is not accurate enough. Therefore, Yu \textit{et al.} \cite{yu2018ds} presented a robust semantic vSLAM for dynamic environments with five threads based on ORB-SLAM2. They used SegNet to segment the movable objects at the pixel level and designed a moving consistency check process to detect the movements of the movable ORB features. Only the semantically and geometrically dynamic features were deleted. Similarly, Cui \textit{et al.} \cite{cui2019sof} combined the results of semantic segmentation from SegNet with ORB-SLAM2. They proposed a new method, called Semantic Optical Flow (SOF), to improve the detection of dynamic features and reasonably remove the dynamic features during tracking. Unlike \cite{zhong2018detect,wang2019computationally,xiao2019dynamic,yu2018ds,cui2019sof} that directly detect and delete the dynamic features, recent studies tried to further estimate and utilize the dynamic objects in the scenes \cite{huang2020clustervo,yang2019cubeslam}. Huang \textit{et al.} \cite{huang2020clustervo} proposed a stereo VO framework that not only estimated the motion of camera but also clustered the surrounding objects. A sliding window optimization was used to solve the motions of camera and surrounding dynamic objects. Yang \textit{et al.} \cite{yang2019cubeslam} dug deeper into the relationship between the motion of camera and surrounding objects, and found that the two parts can improve each other. Since both dynamic and static objects can provide long-range geometric and scale constraint, it is helpful to improve the camera pose estimation and constrain the monocular drift.

\textbf{Scale recovery and visual localization:} Scale ambiguity has always been a big challenge for monocular vSLAM, which brings great uncertainty to accurate trajectory prediction and mapping \cite{taketomi2017visual}. Because objects in reality have their own inherent properties, like the height of cars, these properties can be used for monocular vSLAM to get the absolute scale information of scenes. Therefore, semantic information can be utilized to build a bridge between objects and their properties, and it has shown its effectiveness in monocular vSLAM for scale recovery and assisting localization.
Semantic information introduces the size information of objects in the environment into the vSLAM framework to handle the problem of monocular scale ambiguity.
Frost \textit{et al.} \cite{frost2018recovering} represented objects in the environment as spheres and recovered the scale from the detected objects with a known radius. Similarly, in \cite{sucar2018bayesian}, Sucar \textit{et al.} recovered the scale by setting the prior height of the object (car). A detection method was used to detect this object and compute the height, and the scale was solved by the ratio of the calculated height to the prior height. For localization, Stenborg \textit{et al.} \cite{stenborg2018long} proposed a novel method that locates the camera based on semantically segmented images, which is different from traditional localization methods based on features.
To obtain more accurate localization, Bowman \textit{et al.} \cite{bowman2017probabilistic} first integrated the geometric, semantic, and IMU information into a single optimization framework and then associated scale information with semantic information. Lianos \textit{et al.} \cite{lianos2018vso} utilized the semantic information of the scenes to establish mid-term constraints in the tracking process, thereby reducing the monocular drift in VO.

\textbf{High-level semantic perception:}
Autonomous systems need to be able to perform high-level tasks, while the point cloud maps built by traditional vSLAM cannot fully meet the requirements of these tasks. Therefore, a multi-level understanding of their surroundings is essential. For instance, autonomous vehicles should have an understanding of the areas that are drivable and those that have obstacles. However, the environments modeled by traditional vSLAM methods are represented by point clouds, which only contain the location of the point and cannot provide any high-level information about 3D objects. Although the current metric representation for vSLAM executes some basic tasks, such as localization and path planning, it is still insufficient for some advanced tasks, such as human-robot interaction, 3D object detection, and tracking. Therefore, high-level and expressive representations will play a key role in the perception of autonomous systems. To obtain high-level perception, an object-level environment representation \cite{civera2011towards} was proposed in 2011 by modeling the objects in advance and matching them in a global point cloud map. Salas \textit{et al.} \cite{salas2013slam++} extended this work in \cite{civera2011towards}. They created an object database to store the 3D models generated by Kinectfusion \cite{newcombe2011kinectfusion} and computed the global descriptor of every object model for quick matching based on \cite{drost2010model}. They also demonstrated that object-level mapping is useful for accurate relocalization and loop detection.
Contrary to building the models in advance, Sunderhauf \textit{et al.} \cite{sunderhauf2017meaningful} proposed an online modeling method for generating the point cloud models of objects, along with a novel framework for vSLAM by combining object detection with data association to obtain semantic maps.
However, traditional geometry-based high-level environment perception requires modeling and matching objects in the environment in advance, which leads to the complexity of the whole process, i.e. only some objects can be modeled and recognized in these methods.

In comparison to an object-level maps, pixel-level semantic maps-based on learning systems are more precise because they present the semantic information of each point in the maps . To improve the accuracy of segmentation and semantic mapping, conditional random fields (CRFs) have been widely used in related works. A voxel-CRF model was presented in \cite{kim20133d} to associate the semantic information with 3D geometric structure, and a dense voxel-based map with semantic labels was constructed. For consistent 3D semantic reconstruction, Hermans \textit{et al.} \cite{hermans2014dense} proposed a novel 2D-3D label transfer method based on CRFs and Bayesian updates. Considering the intrinsic relationship between geometry and semantics, Kundu \textit{et al.} \cite{kundu2014joint} utilized the constraints and jointly optimized semantic segmentation with 3D reconstruction based on CRFs. Gan \textit{et al.} \cite{gan2017sparse} focused on the continuity of maps and valid queries at different resolutions, and exploited the sparse Bayesian inference for accurate multi-class classification and dense probabilistic semantic mapping. With the help of semantic maps, autonomous systems can obtain a high-level understanding of their surroundings, and they can easily know ``which and where is the desk''.

With the development in deep neural networks, several detection and segmentation methods based on deep learning are proposed. Methods for object detection and image segmentation have been reviewed in \cite{zhao2019object} and \cite{ghosh2019understanding}. Leveraging deep learning-based image segmentation to perform semantic mapping is also a hot topic. In \cite{li2016semi}, Li \textit{et al.} combined the LSD-SLAM \cite{engel2014lsd} with CNN-based image segmentation to reconstruct a semi-dense semantic map. Cheng \textit{et al.} \cite{cheng2017dense} integrated a CRF-RNN-based segmentation algorithm with ORB-SLAM \cite{mur2015orb}, and built a dense semantic point-cloud map by using RGB-D data. Deep learning-based semantic segmentation with dense SLAM frameworks have also been applied to construct dense semantic maps. Mccormac \textit{et al.} \cite{mccormac2017semanticfusion} incorporated CNN-based semantic prediction into state-of-the-art dense vSLAM method, ElasticFusion \cite{whelan2016elasticfusion}. They considered the multi-view segmentation result of the same 3D point and fuse semantic information in a probabilistic manner.
%Zhang \textit{et al.} \cite{zhang2018semantic} use a filter process to remove the unstable features computed by YOLO.

When/where can we integrate learning methods to aid traditional frameworks, like vSLAM? There are two main ways of using deep learning to improve traditional frameworks: one is to enhance the quality of inputs through the learning systems, like image enhancement and vSLAM; the other is to embed the learning systems into traditional frameworks, like pose estimation, depth estimation and vSLAM. For example, considering that traditional vSLAM cannot well adapt to challenging low-light environments, learning methods are used to enhance the stability of feature tracking of vSLAM by improving the quality of input images \cite{jung2019multi}. Since dynamic objects will affect the feature matching, which in turn affects the pose and depth solution of vSLAM, learning systems are used to detect dynamic objects and help to eliminate the dynamic features \cite{xiao2019dynamic}. Therefore, the basic idea is to analyze the limitations and shortcomings of traditional vSLAM, and then introduce learning systems to improve the traditional vSLAM framework. In addition, we should also note that the introduction of the learning systems also brings some problems to the entire framework, such as the increase of computation, the dataset dependence of the learning systems, etc, and there also remain problems we need to address in the future.

%-------------------------------------------------------------------------
\section{Autonomous visual navigation}

\begin{figure*}[t]
	\centering
	\includegraphics[width=0.8\textwidth,height=0.24\textwidth]{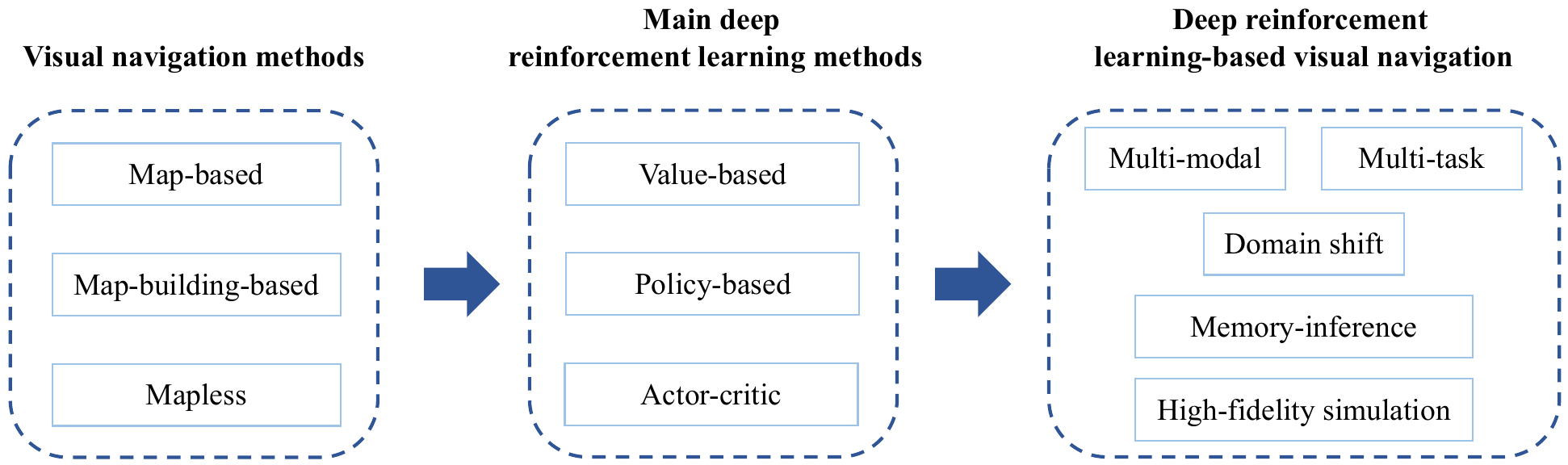}
	%\caption{fig1}
	\caption{The structure of autonomous visual navigation based on reinforcement learning.}
	\label{fig:fig4}
\end{figure*}

After perceiving the surroundings and state, autonomous robots will plan appropriate trajectories according to the missions, their own state as well as the environmental information. A survey of geometry-based motion and control planning for autonomous vehicles is proposed in \cite{paden2016survey}. Therefore, in this section, we mainly focus on autonomous visual navigation based on reinforcement learning, as shown in Fig. \ref{fig:fig4}. We first present visual navigation methods, and introduce three main deep reinforcement learning methods. Then, we review deep reinforcement learning-based visual navigation scenarios, methods and environments.

Navigation can be defined as a process of accurately determining one's location, planning, and following a route from one place to another. With the help of the advanced sensors and navigation algorithms, vision has been introduced into navigation \cite{ma2018nonrigid}, \cite{hu2021vision}. Compared with other navigation methods, such as magnetic navigation \cite{gozick2011magnetic}, inertial navigation \cite{barshan1995inertial}, laser navigation \cite{stahn2007laser} and GPS navigation \cite{skog2009car}, visual navigation has a relatively low cost and general simulation platforms. Therefore, visual navigation has become a mainstream research approach for researchers. Traditional visual navigation of mobile robots is generally based on three main methods: \textrm{map-based navigation, map-building-based navigation, and mapless navigation} \cite{bonin2008visual}.
%Navigation can be defined as a process of accurately determining one's location, planning and following a route from one place to another. With the help of advanced sensors and navigation algorithms, vision has been introduced into navigation \cite{ ma2018nonrigid  }, \cite{ valeiras2018event }. Vision-based navigation leads to wider researches and provides many different solutions in the fields of vision and control. Traditional visual navigation of mobile robots is generally based on three main methods: \textrm{map-based navigation, map-building-based navigation, and mapless navigation} \cite{bonin2008visual}.

Map-based navigation requires the global map of the current environment to make decisions for navigation. For example, in \cite{kim1999symbolic}, the robot used a generic map to accomplish symbolic navigation. Specifically, the robot was not guided to the locations with specific coordinates but with symbolic commands. Symbolic commands are the general description of the types of entities in the environment.
In map-building-based navigation, robots use different sensors to perceive the environment and update the map. For example, in \cite{gaspar2000vision}, the robot accomplished long-distance navigation with the help of a topological map. Specifically, the global environment was built as a topological map and described by graphics during navigation. An appearance-based system and a visual servoing strategy qualitatively estimated the position of the robot and kept it on a specific trajectory employing omnidirectional cameras.
In mapless navigation, robots do not have any environment information and navigate with the perceived information without maps. Saeedi \textit{et al.} \cite{saeedi2006vision} presented a general-purpose 3-D trajectory-tracking system. This system could be applied to unknown indoor and outdoor environments without the need of mapping the scene, odometry or the sensors other than vision sensors.

\textbf{Reinforcement learning based visual navigation:} Since reinforcement learning is suitable for continuous motion planning tasks in complex environments, reinforcement learning based navigation has been preliminarily studied recently. Compared to traditional control methods, when using reinforcement learning algorithms to address navigation problems, sufficient theoretical knowledge is not required, and the proposed model tends to solve the problem end-to-end.
By defining better state space representations in complex and infinite environments, reinforcement learning algorithms can be simplified and navigation efficiency will be improved. Jaradat \textit{et al.} \cite{jaradat2011reinforcement} used Q-learning to handle the problem of mobile robot navigating in an unknown dynamic environment. Owing to the infinite number of states in a dynamic environment, the authors limited the number of states based on a new definition of the state space to ensure that the navigation speed was improved. Similarly, Shi \textit{et al.} \cite{shi2018adaptive} utilized Q-learning to predict partial missing QR codes in order to ensure image-based visual servoing. Since the QR code has a large number of feature points, the authors proposed to take its rotation and translation between the current image and the desired image as the state space to simplify the computational complexity of reinforcement learning.%Bruce \textit{et al.} \cite{bruce2017one} introduced a reinforcement learning method to map-building based navigation. They employed interactive replay for a single traversal of the real environment to build the map and guided a mobile robot to a fixed target in the environment using SLAM techniques without involving humans. In the navigation process, pre-trained visual features and random observations are used to expand the training set. Therefore, the robot can successfully navigate without the need for fine-tuning under environment changes.

During the training period, adding auxiliary tasks, such as value function \cite{ bellemare2019geometric}, reward prediction \cite{jaderberg2016reinforcement}, map reconstruction \cite{mirowski2016learning}, and edge segmentation \cite{li2019deep}, can improve the reinforcement learning efficiency. Jaderberg \textit{et al.} \cite{jaderberg2016reinforcement} proposed a novel unsupervised reinforcement and auxiliary learning algorithm. The algorithm predicted and controlled the features of the sensorimotor stream by treating them as pseudo-rewards for reinforcement learning. Moreover, during the training process, the agent was allowed to perform additional tasks, such as pixel control, reward prediction and value function replay. In \cite{mirowski2016learning}, the agent only used the visual information (images of the monocular camera) for navigation search (finding the apple in the maze). The study considered two auxiliary tasks. In the first task, a low-dimensional depth map was reconstructed at each time step, which is beneficial for obstacle avoidance and short-term path planning. The other task involved loop closure detection, wherein the agent learned to detect whether the current location had been visited within the currently running trajectory. The experiments in these studies proved that co-training can significantly improve the learning speed and performance of the model.

Recently, multi-modal reinforcement learning has become a hot point and cutting edge, which combines multi-modal information, such as language and video, with vision as inputs in the reinforcement learning model. To deal with navigation issues, visual language navigation (VLN) \cite{anderson2018vision} has been widely used in recent years. VLN is a task that guides the embedded agent to execute natural language instructions in a 3D environment. It requires a deep understanding of the linguistic semantics, visual perception, and most importantly, the alignment of the two. Most existing methods are based on sequence-to-sequence architecture \cite{ke2019tactical,zhu2020vision,sun2021depth}. That is, instructions are encoded as word sequences, and navigation trajectories are decoded as a series of actions, which are enhanced by attention mechanism and beam search. Therefore, connecting cross-modality training data is a key to improve training efficiency. Wang \textit{et al.} \cite{wang2019reinforced} summarized VLN tasks and studied how to solve the three key challenges of VLN, namely cross-modal grounding, ill-posed feedback, and the generalization problems. Chaplot \textit{et al.} \cite{chaplot2019embodied} proposed a dual-attention unit to disentangle the knowledge of words in the textual representations and visual concepts in the visual representations, and align them with each other. The fixed alignment enables the learned knowledge transferred across tasks. In response to the first and second challenges, the authors proposed the reinforced cross-modal matching (RCM) method, which used reinforcement learning to connect local and global scenarios. In response to the third challenge, self-supervised imitation learning (SIL) was proposed, which helped the agent to get better policies by imitating its best performance from the past.

However, reinforcement learning-based navigation is limited to small action space and sample space, and it is generally in a discrete situation. Moreover, more complex tasks closer to the actual situation tend to have a large state space and continuous action space.

\textbf{Deep reinforcement learning based visual navigation:} has achieved promising results recently by combining the perceptual ability of deep learning with the continuous decision ability of reinforcement learning. Compared to reinforcement learning based navigation, deep reinforcement learning methods equip robots with the ability to learn high-dimensional data \cite{lin2020blockchain} to ensure precise perception and positioning, so that they can accomplish more complex tasks, for example, navigating to different targets in a scene without retraining \cite{zhu2017target}.
%, as shown in Fig. \ref{fig:fig9}.

Deep reinforcement learning algorithms can be divided into two types: value-based and policy-based. Value-based algorithms learn the value function or the approximation of the value function, and then select a policy based on the value. Deep Q-Network (DQN) is the first value-based algorithm. Tai \textit{et al.} \cite{tai2016robot} first built an exploring policy for robotics based on DQN, in order to explore a corridor environment with the depth information from an RGB-D sensor only.
There are many extensions of DQN in order to improve stability and efficiency during training.
Dueling DQN \cite{wang2016dueling} can directly learn the value of state through the advantage function, which makes it learn faster than DQN when some of the actions do not affect the environment. On the other hand, double DQN \cite{van2016deep} can train two Q networks at the same time and choose a smaller Q value to reduce the overestimation error, which equips double DQN with stable performance. In that way, combining dueling DQN with double DQN is a good choice. Zeng \textit{et al.} \cite{zeng2021simultaneous} utilized dueling double DQN with multi-step learning to handle coverage-aware UAV navigation problem. Specifically, the signal measured on the UAV was used to directly train the action-value function of the navigation policy, thus greatly maintaining the relative stability of the target and improving the learning efficiency.
The original DQN can only be applied in tasks with a discrete action space. In order to extend to continuous control, many policy-based algorithms have been developed. Policy-based algorithms learn directly based on the policy without the reward. For example, deep deterministic policy gradients (DDPG) \cite{lillicrap2015continuous} and normalized advantage function (NAF) \cite{gu2016continuous} are policy-based algorithms that have been widely used. In comparison to NAF, DDPG needs less training parameters. Liu \textit{et al.} \cite{liu2018energy} navigated a group of agents to provide long-term communications coverage, which only used one agent to output control decisions for all agents by employing DDPG. However, the DDPG algorithm requires researchers to spend a lot of time iterating and manually adjusting rewards in practice. To address this problem,
%Chiang \textit{et al.} \cite{chiang2019learning} used AutoRL, an evolutionary automation layer around reinforcement learning, which helped to optimize the reward and the neural network architecture. Specifically, AutoRL simultaneously trained a group of agents using DDPG for several generations. Each agent had a slightly different reward function to optimize the real goal-reaching the destination. However, the DDPG algorithm is time-consuming in searching for hyperparameter and manually adjusting rewards in practice.
one way is to use some expansion of DDPG to improve sampling efficiency \cite{tai2017virtual, zhang2019asynchronous}.
Tai \textit{et al.} \cite{tai2017virtual} presented a model using asynchronous multithreading DDPG to collect data, which helped to improve sampling efficiency. The mapless motion planner could be trained end-to-end without any features designed by human or prior demonstrations. Similarly, Zhang \textit{et al.} \cite{zhang2019asynchronous} proposed asynchronous episodic DDPG, which improved learning efficiency with less training time in computationally complex environments. Episodic control and a novel type of noise were introduced to the asynchronous framework in order to improve sample efficiency while increasing data throughput.
Another solution is to introduce AutoRL, an evolutionary automation layer around reinforcement learning, which helps to optimize the reward and the neural network hyperparameter while learning navigation policies.
Chiang \textit{et al.} \cite{chiang2019learning} introduced AutoRL to simultaneously train a group of agents using DDPG for several generations. Each agent had a slightly different reward function and hyperparameter to optimize the real goal-reaching the destination.

Actor-critic (AC) algorithm \cite{ konda2000actor } combines two types of deep reinforcement learning algorithms mentioned above. That is, the actor network chooses the proper action in a continuous action space, while the critic network implements single-step-update, which improves the learning efficiency. In other words, it learns both the value function and the policy function. Asynchronous Advantage Actor-Critic (A3C) network \cite{mnih2016asynchronous}, an improvement of the AC network with multithreading method, as an on-policy learning algorithm, uses newly collected samples for each gradient step.
A3C network performs interactive learning with the environment in multiple threads simultaneously, thereby avoiding over-fitting of the training data. When robots autonomously exploring unknown cluttered environments, A3C equips the robots with the ability to gain cross-target generalization.
In order to gain cross-target generalization ability, Zhu \textit{et al.} \cite{zhu2017target} took both the target and scene images as inputs of the deep reinforcement learning network; then, the agent followed the output action to navigate to a target. During the training process, a new observation was valued through the A3C network to ensure that the agent did not need to retrain the new target.
Moreover, Duron \textit{et al.} \cite{druon2020visual} added semantic network to the visual network proposed in \cite{zhu2017target} to learn context from the objects present in the scene.
A3C network takes the features from the joint embedding layer as inputs and then outputted the next action and the Q-value for the current state.
Besides, off-policy learning algorithm, such as soft actor-critic (SAC) \cite{haarnoja2018soft}, aims to reuse past experience, which provides for both sample-efficient learning and stability. Jesus \textit{et al.} \cite{de2021soft} applied SAC to learn continuous action space policies and maximize the entropy of the policy in the mobile robotics exploration problem.

With the development in deep reinforcement learning algorithms, the problem of vanishing gradient arises. That is, as the number of hidden layers in neural networks increases, the classification accuracy in the training process decreases. The LSTM architectures \cite{lee2019analysis} is a good way to tackle this problem. When the input data is time-varying, LSTM can capture the long-term dependencies of sequential data. Mnih \textit{et al.} \cite{mnih2016asynchronous} used LSTM units to make better decisions by considering the previous state characteristics.
%With the same purpose, in \cite{mirowski2016learning}, the authors combined LSTM with auxiliary tasks and A3C networks to navigate in a maze to find a target using the images of the monocular camera.
In real-word navigation, training data are more variable and unpredictable than those in simulation experiments. Therefore, LSTM plays a vital role in generating good navigation policies. Mirowski \textit{et al.} \cite{mirowski2018learning} only used the visual information as input for unmanned vehicle navigation without relying on maps, GPS, and other auxiliary tools. The authors put unmanned vehicles in complex scenes of city scale and collected real-world data for training. To accomplish the tasks, a multi-city navigation network with LSTM was proposed. The method processed images, extracted features, remembered and understood the environment, and finally generated the navigation policies.

By using deep reinforcement learning methods, agents can automatically learn the characteristics of the data collected by the sensors without human intervention. On this basis, agents are able to formulate a navigation policy to ensure navigation in more complex environments, especially in real world. In the field of navigation that is biased toward obstacle avoidance, the methods used in \cite{chen2018deep,xie2017towards} obtained satisfactory generalization performance. Therefore, the models trained solely in virtual environments are possible to be transferred to real robots. Chen \textit{et al.} \cite{chen2018deep} presented a novel approach to train action policies to acquire navigation skills for wheel-legged robots using deep reinforcement learning. It is crucial that domain randomization was introduced to increase the diversity in training samples, improve the generalization ability, thereby focusing on the task-related aspects of observation. Therefore, it has been used in real environments with more complicated types of obstacles and movements. Xie \textit{et al.} \cite{xie2017towards} proposed a new network structure, consisting of two parts, to deal with the obstacle avoidance problems. First, the convolutional residual network was used to extract the depth information. Then the reinforcement learning structure could efficiently learn how to avoid obstacles in a simulator even with very noisy depth information predicted from the RGB images.

To improve the performance of deep reinforcement learning networks, training data should be essentially considered in experiments. Sufficient and variable training data are the basis of convincing results during the training process, while in the real-world, training data are always unobtainable or missing. To handle this problem, simulation frameworks can be utilized to train agents. For example, in \cite{zhu2017target}, the first simulation framework, called AI2-THOR (The House Of inteRactions), was developed to provide an environment with high-quality 3D scenes as well as physics engines. Therefore, the robot in a simulation environment can effectively collect several training samples, which improves the data utilization.
Specifically, Wu \textit{et al.} \cite{wu2021reinforcement} analyzed the cross-target and cross-scene generalization ability of the target-driven navigation models on AI2-THOR. The evaluation, which was conducted in 120 synthetic scenes from four categories, including kitchen, living room, bedroom and bathroom, greatly exceeded some relative baselines.

After training in simulation, it is difficult to ensure that the agent achieves similar performances between the virtual scene and the real scene because of the domain shift. One possible solution is to add the vSLAM map in the navigation process, which helps to narrow the difference in performance between simulation and real environment. On the basis of \cite{chiang2019learning}, Francis \textit{et al.} \cite{francis2020long} introduced the vSLAM map to robot navigation, in order to reconstruct the motion probability map. Since the vSLAM map is noisy, it can compensate for the difference in performance between the robots in the virtual and the real environment due to the different levels of noise. From another perspective, constructing an exploration framework bridges the gap between simulation and the real environment. Li \textit{et al.} \cite{li2019deep} constructed a framework consisting of mapping, decision and planning. Each module was independent and can be achieved by a variety of methods. Compared to traditional end-to-end deep reinforcement learning methods with raw sensor data as input and control policy as output, the proposed deep reinforcement learning algorithm based on framework learned faster and equipped itself with better generalization performance in different maps.
%-------------------------------------------------------------------------

\section{Discussion}

\subsection{Deep learning-based visual perception}

The constructed map is an intuitive representation of the scene perception and the basis for intelligent robots to autonomously perform advanced tasks. Mapping has undergone a development process \textit{from 2D to 3D, from sparse to dense, and from topological to semantic, among others}. Furthermore, although several methods have been proposed to improve the localization accuracy, there are still many challenges remaining to be solved. Therefore, we summarize the challenges and promising directions of perception as follows.
\begin{itemize}
	\item \textbf{Accurate perception:} Although learning-based perception algorithms have made great process in the perception areas, their accuracy, especially the accuracy of unsupervised learning methods, still has much room for improvement. Digging the more effective constraints for training from the aspect of geometry, cross-task relationships and interpretability, utilizing novel learning frameworks, like meta-learning, curriculum learning and lifelong learning to make full use of the data, and developing more efficient neural network frameworks for feature extraction and inference are both promising directions.
	\item \textbf{Robust perception:} Robustness is one of the most important indicators for the reality application of perception algorithms. Although current learning systems have received good accuracy on the datasets, the network will be affected by the \textit{sensor noise, lighting and scenarios} when being used to real environments. Therefore, the robust environmental perception, ego-motion perception and navigation based on learning systems under different conditions (like  \textit{different seasons, different weather, different lighting conditions, different source sensors, indoor and outdoor as well as day and night}) in the same scene are problems to be handled.
	\item \textbf{Real-time perception:} Real-time perception is important for autonomous systems in practical applications. Current high accuracy networks are based on complex network structures, which includes a huge number of parameters and large Flops. Therefore, the training and application of deep neural networks have a higher demand on the \textit{computing power} of the systems, which limits the practical applications. Using novel lite-weight learning architectures, such as \textit{light weight network and knowledge distillation}, to improve the real-time performance of perception networks will be another trend.
	\item \textbf{Geometry assist in perception:} Utilizing \textit{the geometric prior} built by a learning framework or knowledge graph in the perception of autonomous systems is helpful and a promising direction with broad development prospects. For example, semantic labels predicted by deep learning are used to correlate with the \textit{knowledge graph} of objects to obtain prior geometric information, such as the size of objects; therefore, the detailed scale, structure, and 3D information can be obtained.	
\item \textbf{Representation of the environment based on deep learning:} \textit{Representing the environment based on deep learning} is another challenge and a promising direction. Although previous works such as \cite{tateno2017cnn,yang2018deep,loo2019cnn} leveraged the deep learning into mapping, the maps of these methods are still built traditionally. With the developments of Nerf algorithms \cite{dellaert2021neural,mildenhall2020nerf}, it provides a way to present the scene by using neural networks. Most recent work has tried to construct the SLAM systems based on Nerf \cite{sucar2021imap}, and this is quite an interesting and promising direction.
	\item \textbf{Multi-sensor data fusion based on deep learning:} Fusing information from \textit{multi-sensors (IMU, LiDAR, event-based camera, or infrared camera )} or multi-agent is an effective way to deal with poor quality input images comprising motion blur and recover scale information. However, expressing the additional sensor information explicitly in the constraints for training is a significant challenge. For example, the current methods leverage IMU data with images for pose estimation in a supervised manner \cite{clark2017vinet,chen2019selective}, and the information from IMU is not represented in the loss function. Thus, whether the IMU data plays an important role in pose estimation and what role it plays is unknown and not yet explainable.
	\item \textbf{Integration of deep learning and traditional frameworks:} Although a lot of relevant research has been summarized above, there is still a lot of work to be done in this direction. With the help of deep learning, the basic idea is to improve the traditional frameworks by analyzing the \textit{limitations and shortcomings} of traditional methods. For example, considering that the current direct methods rely heavily on the photometric consistency assumption, we can use deep learning to perform a \textit{photometric correction} or transform images into \textit{photometric-consistent feature maps}.
\end{itemize}

\subsection{Reinforcement learning based visual navigation}

%\textbf{Reinforcement learning:}
There is still a long way to go before reinforcement learning can be applied to autonomous systems.Therefore, there are many challenges to be addressed.
\begin{itemize}
	\item \textbf{Sparse rewards: } Rewards have a great impact on the learning results during the training process, but the problem of \textit{sparse rewards} in reinforcement learning has not been well solved. When the training tasks are complicated, the probability of exploring the target (getting positive rewards) by random exploring becomes very low. Therefore, it is difficult for reinforcement learning algorithms to converge by only relying on the positive rewards. To deal with this problem, redesigning the reward function according to specific scenarios will be helpful in avoiding the problem of \textit{sparse rewards}, and by means of hierarchical networks, such as \textit{hierarchical reinforcement learning} \cite{pateria2021hierarchical}, the training efficiency and the final performance will be improved.
	
	\item \textbf{Complicated calculation and high cost: } When a robot navigates in a large-scale or continuous state space environment, the calculation process is likely to be high-dimensional and complex. Moreover, reinforcement learning algorithms require thousands of trials and errors to train iteratively, while in real-world tasks, agents hardly withstand so many trials and errors because of limited cost. Therefore, the efficient feature representation of large-scale spatial reinforcement learning will simplify the calculation process. In other words, the use of efficient and fast online self-evaluation reinforcement learning algorithms will reduce the training costs and improve learning efficiency.

	\item \textbf{Performance between simulation and real-world:} Due to the huge gap between simulation environments and real scenes, many reinforcement learning algorithms with higher performance in simulation experiments cannot handle the practical problems in the real world, which strongly limits the widespread application of this technology. Establishing a network that can be directly transferred to the real world and building a high-fidelity simulation and physical platform will be a future trend, which can effectively convert virtual scenes generated in a simulator into real scenes for reinforcement learning training.

	\item \textbf{Transferable property improvement:} In many tasks, the training data is limited and unobtainable. \textit{Adversarial learning} methods \cite{zhang2020autonomous} can be applied to increase the data differences in the training process and reduce data differences in the testing process, which improves the data diversity and the generalization ability of the model. Moreover, many transfer learning methods, such as \textit{few-shot learning, zero-shot learning, and meta learning} \cite{wang2020generalizing, wang2019survey, hospedales2020meta}, can recognize the new model and apply the knowledge and skills learned in previous tasks to novel tasks with few or even no training data. This is effective for enhancing the transferability, reducing network parameters, and promoting generalization.

	\item \textbf{Multi-modal and multi-task:} Current reinforcement learning-based navigation methods mainly focus on visual input. However, by considering the information from multiple models, such as voice, text, and video, the agents can better understand the scenes and the performance in experiments will be more accurate and efficient. Moreover, it is proved that \textit{multi-task reinforcement learning} models \cite{zeng2021decentralized}, in which the agent is simultaneously trained with auxiliary and target tasks, improve the training efficiency. Therefore, \textit{multi-model and multi-task} are also development trends in navigation based on reinforcement learning.

\end{itemize}

\subsection{Application}

The developments of autonomous environment perception and motion planning drive the emergence of a large number of high-tech industries, such as unmanned vehicles and service robots, which have greatly improved the quality of human life \cite{severinson2003social}. Furthermore, autonomous systems have a broad application prospect in various fields, like industry, agriculture, services, transportation, etc. For example, accidents in petrochemical industry have occurred from time to time in recent years, which inevitably caused great damage of life and property. The use of autonomous systems monitoring a chemical park can help to find dangers in advance. Intelligent monitoring robots with various gas and optical sensors can monitor the safety hazards in the chemical plant area in real time. Robots autonomously perceive and construct the map of structure environments based on visual sensors. Then, based on the perceived information, robots plan the path and tasks for better monitoring. In case of emergency, the environment becomes semi-structured and complicated, in which autonomous robots can reach dangerous areas, sense the surrounding areas, deliver important information to the staff, assess the situation and even assist staff in decision-making as well as rescue. At present, the difficulties lie in the distance between theoretical research and practical applications, such as reliability, robustness and real-time response capability. Therefore, this survey reviews the existing learning-based perception and navigation methods, which provides a guideline for future research and promotes the developments of autonomous systems.

%-------------------------------------------------------------------------
\section{Conclusion}

 Through this review, we aim to contribute to this growing area of research by exploring the learning-related methods for self-state perception, environment perception and navigation in autonomous systems.
Therefore, we review the related works of learning-based vSLAM and navigation in the learning age.
The influx of deep learning algorithms can be observed to support the subtasks of vSLAM or incorporate with vSLAM in recent works, which improve the robustness and performance of traditional vSLAM algorithms. Meanwhile, navigation based on deep reinforcement learning achieves good efficiency and transferability in autonomous systems. We provide two comprehensive taxonomy tables of state-of-the-art vSLAM algorithms as well as deep learning-based depth and pose estimation methods, which clarify the mainstream algorithm framework and the development trend.
Finally, this review highlights the key challenges and promising directions in learning-based perception and navigation.

%%%%%%%%%%%%%%%%%%%%%%%%%%%%%%%%%%%%%%%%%%%%%%%%%%%%%%%
%\section*{ACKNOWLEDGMENT}

% or




%\bibliographystyle{IEEEtran}
%\begin{thebibliography}{1}

{%\small
	\bibliographystyle{IEEEtran}
	\bibliography{egbib}
}
%\end{thebibliography}

\newpage


%\end{IEEEbiography}


%\newpage



%\vfill


\end{document}